\documentclass[lettersize,anonymous,journal]{IEEEtran}
\usepackage{amsmath,amsfonts}
\usepackage{algorithmic}
\usepackage{algorithm}
\usepackage{array}
\usepackage[caption=false,font=normalsize,labelfont=sf,textfont=sf]{subfig}
\usepackage{textcomp}
\usepackage{stfloats}
\usepackage{url}
\usepackage{verbatim}
\usepackage{graphicx}
\usepackage{cite}
\usepackage{booktabs}
\usepackage[dvipsnames]{xcolor}
\begin{document}

\title{GSwap: Realistic Head Swapping with Dynamic Neural Gaussian Field}

\author{Jingtao~Zhou,
        Xuan~Gao,
        Dongyu~Liu,
        Junhui~Hou,
        Yudong~Guo$^\dagger$,
        Juyong~Zhang% <-this % stops a space
% \IEEEcompsocitemizethanks{\IEEEcompsocthanksitem X. Gao, D. Liu, Y. Guo, % and J. Zhang are with the School of Mathematical Science, University of % Science and Technology of China (USTC); J. Hou is with the Department of % Computer Science, City University of Hong Kong (CityU); J. Zhou is with % both USTC and CityU.
% % note need leading \protect in front of \\ to get a newline within % \thanks as
% % \\ is fragile and will error, could use \hfil\break instead.
% }% <-this % stops an unwanted space
\thanks{J. Zhou is with the School of Mathematical Science, University of Science and Technology of China and the Department of Computer Science, City University of Hong Kong.}
\thanks{X. Gao, D. Liu, Y. Guo, and J. Zhang are with the School of Mathematical Science, University of Science and Technology of China.}
\thanks{J. Hou is with the Department of Computer Science, City University of Hong Kong.}
\thanks{$^\dagger$Corresponding author. Email: \texttt{yudong@ustc.edu.cn}.}}

\markboth{IEEE Transactions on Visualization and Computer Graphics}%
{Shell \MakeLowercase{\textit{et al.}}: Bare Demo of IEEEtran.cls for Computer Society Journals}


\maketitle

\begin{abstract}
% We present GSwap, a consistent and realistic video head swapping system empowered by dynamic neural Gaussian portrait priors. Previous methods mainly use 2D generative models or 3DMM face models for face swapping, which often suffer from poor 3D consistency and limited quality. Moreover, these approaches face challenges in full head-swapping tasks due to the lack of comprehensive head modeling and effective background blending. Our approach addresses these limitations by embedding an intrinsic 3D Gaussian feature field within a full-body SMPL-X surface, elevating 2D portrait videos to a dynamic neural Gaussian field. This guarantees 3D consistent portrait rendering and natural head-torso relationship. To create the training dataset for GSwap, we adapt a pretrained 2D portrait generative model to the source head domain with few-shot source images. Additionally, we introduce a neural rerendering strategy that seamlessly integrates foreground and background. Results demonstrate that our method outperforms previous approaches in quality, naturalness, 3D consistency, and identity preservation.

We present GSwap, a novel consistent and realistic video head-swapping system empowered by dynamic neural Gaussian portrait priors, which significantly advances the state of the art in face and head replacement. Unlike previous methods that rely primarily on 2D generative models or 3D Morphable Face Models (3DMM), our approach overcomes their inherent limitations, including poor 3D consistency, unnatural facial expressions, and restricted synthesis quality. Moreover, existing techniques struggle with full head-swapping tasks due to insufficient holistic head modeling and ineffective background blending, often resulting in visible artifacts and misalignments. To address these challenges, GSwap introduces an intrinsic 3D Gaussian feature field embedded within a full-body SMPL-X surface, effectively elevating 2D portrait videos into a dynamic neural Gaussian field. This innovation ensures high-fidelity, 3D-consistent portrait rendering while preserving natural head-torso relationships and seamless motion dynamics. To facilitate training, we adapt a pretrained 2D portrait generative model to the source head domain using only a few reference images, enabling efficient domain adaptation. Furthermore, we propose a neural re-rendering strategy that harmoniously integrates the synthesized foreground with the original background, eliminating blending artifacts and enhancing realism. Extensive experiments demonstrate that GSwap surpasses existing methods in multiple aspects, including visual quality, temporal coherence, identity preservation, and 3D consistency.

\end{abstract}


\IEEEdisplaynontitleabstractindextext
\begin{figure*}[h]
    \centering
    % \fbox{\rule{0pt}{2in} \rule{.9\linewidth}{0pt}}
    \includegraphics[page=1,width=\textwidth]{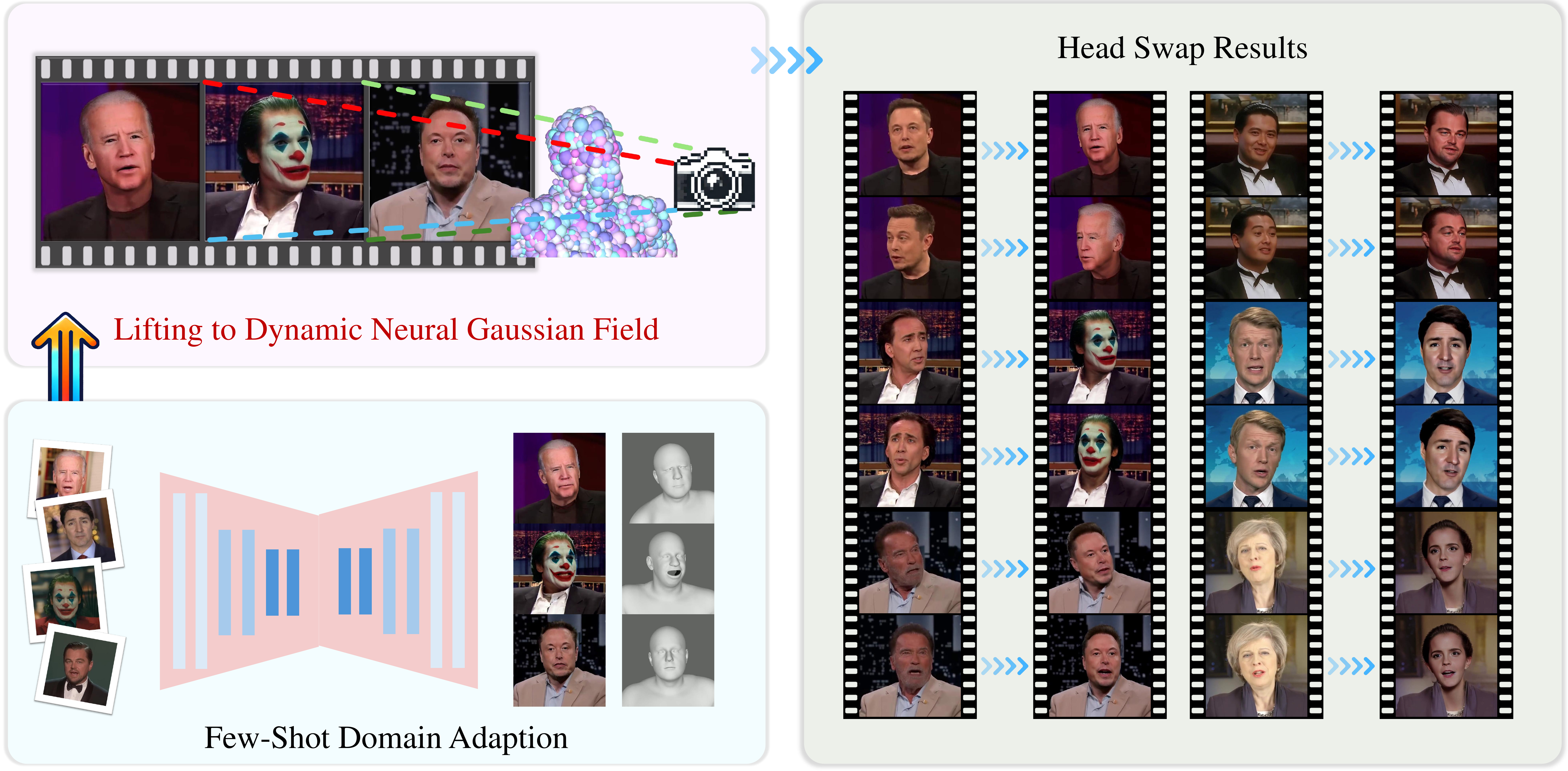}
    \vspace{-5mm}
    \caption{Given a target video and few-shot source images, our GSwap firstly adapts a pre-trained 2D portrait generative model to the source head domain. It then elevates the 2D video head swapping task into the dynamic neural gaussian field, significantly enhancing the temporal consistency and overall quality of the head swap.}
    \label{fig:teaser}
    \vspace{-3mm}
\end{figure*}
\IEEEpeerreviewmaketitle

\section{Introduction}
% Video Head Swapping, the ability to seamlessly replace the head in a target video with a source one, has extensive applications in various entertainment scenarios such as film, art, and AR/VR. Ensuring the identity similarity and temporal consistency while filling the region mismatch between swapped head edges and backgrounds make this task to be challenging.

\IEEEPARstart{H}{ead} swapping has extensive applications in fields such as film, art, and AR/VR. How to seamlessly transfer the identity of the source subject to the target video while keeping the attributes (e.g., pose, expression, background) unchanged remains a very challenging problem.

% Early works mainly focus on face swapping task\cite{chen2020simswap,li2019fs,Nirkin2019fsgan,Perov2020deepfacelab,Zhu_2021_CVPR}, most of them consist of two stages, face reenactment and image blending. However, besides facial identity and expression modeling, Head swapping also requires the structural information of a whole head and the non-rigid hair. In addition to this, the region mismatch between swapped head edges and backgrounds caused by the editing of head shapes and hairstyles is a difficult problem for head swapping. Morevoer, the color difference between source and target skins needs to be handled carefully. Deepfacelab \cite{Perov2020deepfacelab} is the first work to tackle head swapping. However, it requires huge amounts of source data are required and the regions that require inpainting the fusion look unnatural. Heser\cite{shu2022head} introduces a GAN-based aligner to extract the whole head information and designs a blender to handle facial skin color and background texture mismatches. However, it is a 2D-awared method that hard to keep 3D consistency. Its quality is limited by the representation ability of the GAN model.

Face swapping task has been an active research topic in recent years. Early methods\cite{chen2020simswap,li2019fs,Nirkin2019fsgan,Perov2020deepfacelab,Zhu_2021_CVPR} utilize GAN models to fuse the identity information from the source face to the target features, employing adversarial training~\cite{goodfellow2014generative} to ensure the authenticity of the face. This kind of methods suffer from the limited representation ability of 2D GAN models. Recent methods adopt diffusion models for face swapping~\cite{zhao2023diffswap,baliah2024realistic}. While these models show promising results for single image, they struggle to maintain temporal consistency in video applications due to the stochastic nature of diffusion sampling. There also exist some methods leveraging the 3D morphable model (3DMM)~\cite{blanz1999morphable} for face swapping~\cite{nirkin2018face,li2021faceinpainter,ijcai2021p0157}. However, 3DMM only models the facial region, posing challenges for extending the technique to full head swapping.

Head swapping is more challenging than face swapping as it not only requires preserving facial features and expressions similar to the target video, but also needs to capture the structural details of the source subject’s head, hair, and neck. Furthermore, the region mismatch between the swapped head and background is difficult to handle. Heser\cite{shu2022head} attempted to address this by blending a reenacted source head with a segmented target video. However, their method struggles with large pose variations and encounters issues with temporal continuity, primarily due to the absence of 3D prior knowledge. Moreover, due to the limited capacity of the GAN model, the swapped results are not satisfying enough. 

In this paper, we propose a video head swap system that: (1) keeps high identity similarity with source images, (2) preserves 3D consistency, (3) has a natural head-torso relationship, and (4) fills the mismatch region naturally. Unlike previous works that focus solely on the 2D domain, we lift the portrait video editing problem into 4D to ensure 3D and temporal coherence. We adapt a pretrained high-quality portrait generation model with few-shot images of the source head and distill the personalized subject knowledge to our neural 3D Gaussian portrait representation. We further propose a neural rerendering module to blend the foreground feature map with the original video frame seamlessly.

Specifically, We first develop a series of fine-tuning strategies to adapt a pretrained 2D portrait generative model~\cite{paraperas2024arc2face} to the source head domain, and then generate a training dataset through diffusion inpainting as proposed in ~\cite{rombach2022highresolution}. In order to ensure 3D and temporal consistency, and to model the head-neck relationship more naturally, we embed an intrinsic 3D Gaussian feature field in a full-body human SMPL-X surface ~\cite{SMPL-X:2019}. We combine the foreground feature map, created by Gaussian splatting, with the background features. Then our neural rerendering technique seamlessly blends these two feature maps. Since the training data, generated by the fine-tuned diffusion model, displays a level of randomness with slight variations in poses and expressions compared to the target video, we re-extract SMPL-X parameters from the training data and introduce a time-conditioned feature to mitigate these inconsistencies. Experiments demonstrate that our approach outperforms existing methods in terms of quality, naturalness, 3D consistency, and identity preservation.

In summary, the main contributions of this paper include:
\begin{itemize}
\item We propose a novel head swap system that maintains consistent identity and expressions. By lifting a 2D video to a dynamic neural Gaussian field based on full body priors, we ensure the 3D and temporal consistency of the head swap result.
\item We distill the knowledge of an adapted portrait generative model into the construction process for identity-preserving swap. To address the inconsistency of the edited dataset, we employ retracking alongside temporal condition features to model the variations across frames.
\item Our neural rerendering module effectively combines foreground and background feature maps, ensuring seamless integration and naturalistic blending in the final output.
% \item Our front feature map and background feature fusion strategy combines rendering and blend while ensuring the natural fit of the portrait foreground and background.
\end{itemize}
\section{Related Work}

\noindent\textbf{Face and Head Swap.}\label{rel-face-head-swapping}
Many methods have been proposed for face swapping, which aims to transfer identity from the source and non-identity attributes from the target to obtain the swapped face. In the classic methods\cite{face2face}, 3DMM~\cite{blanz1999morphable} is utilized to conduct face swapping. However, these methods suffer from unnatural results, and blending constraints the change in face shape limited by 3DMM's capability.
Some methods\cite{Perov2020deepfacelab, Nirkin2019fsgan, liu2022fine, naruniec2020, Otto2022} utilize a reenact model to transfer the identity and blend it on the target body. These methods can hardly handle the difficulties in merging non-identity attributes.
Some methods\cite{li2020advancing, chen2020simswap, ijcai2021p0157, Gao_2021_CVPR, Kim_2022_CVPR, li2024ivafs, luo2024codeswap} apply one encoder to extract identities and another encoder to extract non-identity attributes. However, limited by the capability of the encoder, these methods can not preserve the identity information very well. Some methods\cite{ zhu2021megafs, xu2022high, xu2022rafswap, liu2022fine, Liu2024AmazingFT, IAASAPFE} utilize GAN models to model face distribution. These methods require a complex disentanglement mechanism to achieve attribute transfer, which still cannot get satisfactory results. Recently, some methods\cite{zhao2023diffswap, zhu2024stableswap, jimaging10010021, yu2024fuseanypart, baliah2024realistic} utilize diffusion model to extract face information. Face swapping methods can not handle hairstyles and head shape differences, which limits the overall similarity between the generated results and the source.

While face swapping has long been a topic of interest, only a few studies have been carried out on the task of head swapping. Deepfacelab \cite{Perov2020deepfacelab} is the first work to tackle head swapping. However, it requires huge amounts of source data and can not inpaint the mismatch region between head and background naturally. Heser\cite{shu2022head} is one of the earliest methods to achieve few-shot head swapping in the wild. However, limited by the capacity of the GAN model, its swapped results are not satisfying enough.

\noindent\textbf{Digital Portrait Representation.}\label{rel-3d-representation}
3DMM~\cite{blanz1999morphable} embeds 3D head shape into several low-dimensional PCA spaces. To improve its representation ability, some work extends it to multilinear models~\cite{cao2013facewarehouse,vlasic2006face}, and non-linear models~\cite{tran2018nonlinear,guo20213d}, articulated models~\cite{li2017learning,Zheng2023pointavatar}. These advancements have facilitated various applications, yet they often struggle to generate photo-realistic results due to inherent limitations in representational capacity.

Implicit representations have been widely used in 3D portrait modeling. Many works propose NeRF~\cite{mildenhall2020nerf} or 3D GAN models to model human head distribution.~\cite{chan2022efficient,deng2022gram,sun2023next3d, egavatar}. There are also works focusing on building parametric head models~\cite{hong2021headnerf,Gafni_2021_CVPR} or blendshape head models~\cite{Gao2022nerfblendshape,xu2023avatarmav,bai2024efficient}. Although implicit representations could achieve satisfied rendering quality, they suffer from limited rendering efficiency.

More recently, 3D Gaussian Splatting (3DGS)~\cite{kerbl3Dgaussians} has been utilized in digital head modeling, yielding notable improvements in both efficiency~\cite{xiang2024flashavatar,dhamo2023headgas,gaussianblendshape} and fidelity~\cite{qian2023gaussianavatars,wang2024gaussianhead,xu2023gaussian}. 
~\\

\noindent\textbf{Diffusion-Based Portrait Generation.}\label{rel-diffusion-generation}
Denoising Diffusion Models~\cite{ho2020denoising} have emerged as a powerful paradigm for generative tasks in computer vision. While initial breakthroughs primarily focused on single-image generation~\cite{paraperas2024arc2face,han2024face,wang2024instantid} and editing~\cite{brooks2022instructpix2pix,zhao2023diffswap}, subsequent advances have expanded their applications to dynamic content creation. Notably, FADM~\cite{FADM} pioneered the use of diffusion models for portrait animation, inspiring a wave of follow-up research. Recent innovations in this direction can be categorized into several technical strands: (1) attention-based reference injection mechanisms, (2) cross-identity training with synthetic data pairs~\cite{Xportrait,yang2024megactor}, (3) geometric-conditioned approaches using landmarks, normal maps, or depth cues~\cite{prinzler2024jokerconditional3dhead,ma2024followyouremoji,yang2025showmaker}, and (4) multi-modal control frameworks for enhanced realism~\cite{zhu2024champ,guan2024talkact}. Alternative formulations have also explored diffusion processes in structured representation spaces, including UV coordinates~\cite{lan2023gaussian3diff} and tri-plane features~\cite{wang2023rodin}.

The main challenge of dynamic portrait generation is to keep the consistency between different frames. Some methods~\cite{wang2023zeroshot,tokenflow2023,zhang2024towards} try modifying the latent space of the diffusion model and introducing the cross-frame attention mechanism to enhance the consistency of the generated results. There also exist some works leveraging 2D optical flows to keep detailed correspondence~\cite{yang2023rerender,ouyang2023codef}. However, due to the lack of 3D modeling, they may suffer from poor 3D and temporal consistency. Furthermore, they may struggle to model complex facial motions and the head-torso relationship in the absence of any 3D human priors. Recent works lift 2D video into 4D field~\cite{shao2023control4d,Gao2024PortraitGen}. These works mainly focus on cross-modal editing. Compared with them, our head swapping task has higher requests in quality, temporal consistency, and identity preservation.  

\begin{figure*}[t]
  \centering
  \includegraphics[width=\linewidth]{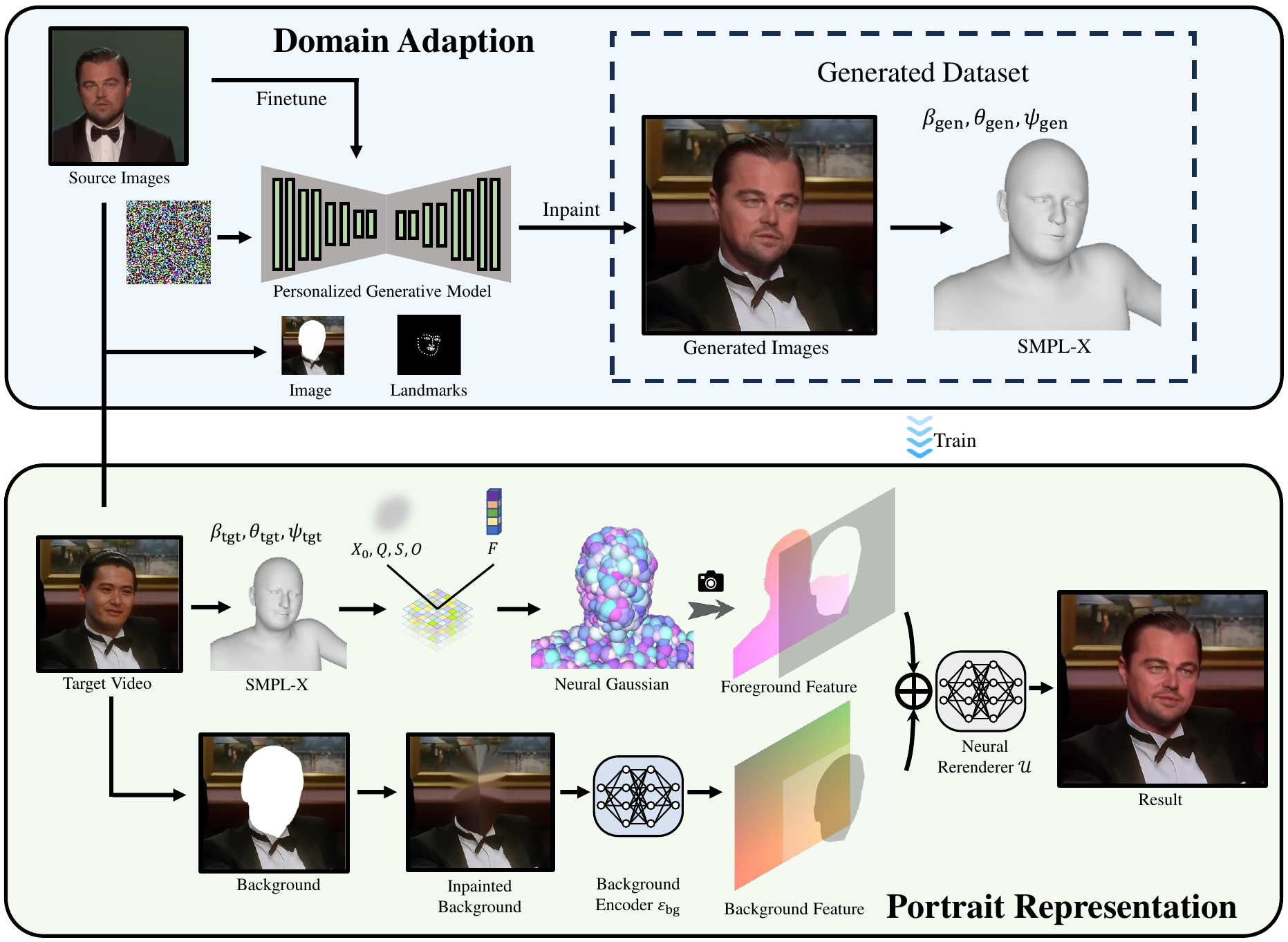}
  \caption{The overall pipeline of our GSwap. Firstly, we adapt a pretrained 2D portrait generation model to the source head domain with source images $\mathcal{I}_{\textrm{src}}$ as input, generating a batch of training data through inpainting (detailed in Fig.~\ref{fig:sec3.1}). Secondly, we introduce the neural Gaussian field as our portrait representation and use a neural rerenderer to handle the mismatch region between foreground and background, generating the output in photo-realistic quality.}
  \label{fig:pipeline}
  \vspace{-2mm}
\end{figure*}

\section{Method}
We propose a consistent and realistic video head-swapping system empowered by dynamic neural Gaussian portrait priors, illustrated in Fig.~\ref{fig:pipeline}. Given few-shot source reference images with $S$ frames, denoted as $ \mathcal{I}_{\textrm{src}}=\{I^{s}_{\textrm{src}}|s \in [1, S]\}$, and a target video with length $T$, denoted as $\mathcal{I}_{\textrm{tgt}}= \{I^{t}_{\textrm{tgt}}|t \in [1, T]\}$. 
Our goal is to generate a head swapped video $\mathcal{I}_{\textrm{out}}=\{I^{t}_{\textrm{out}}|t \in [1, T]\}$, where each output frame $I_{\textrm{out}}^{t}$ aligns with the corresponding target frame $I_{\textrm{tgt}}^{t}$ in terms of pose and expression, while retaining the identity, skin color, head structure, and hairstyle of the source images in $\mathcal{I}_{\textrm{src}}$. To simplify the notation, we will henceforth omit the frame indices $s$ and $t$ in our further descriptions.
% We aim to create head swapped video $\mathcal{I}_{\textrm{out}}=\{I^{t}_{\textrm{out}}|t \in [1, T]\}$. The pose and expression of $I_{\textrm{out}}^{t}$ should remain consistent with the corresponding frame of $I_{\textrm{tgt}}^{t}$, the identity, skin color, head structure and hair style should be the same as $\mathcal{I}_{\textrm{src}}$. For convenience of notation, we omit the frame index subscript $s,t$ in following description.
We first design a series of finetuning strategies to adapt a pretrained 2D portrait generation model to the source head domain, generating a batch of training data through inpainting (Sec.~\ref{arch-gen}). Then we use the generated data frames to supervise the dynamic neural Gaussian portrait model. To handle the inconsistency of generated training data frames, we employ SMPL-X retracking techniques. Additionally, we incorporate temporal conditional features within the Neural Gaussian Texture mechanism to effectively handle the inconsistency across different frames. (Sec.~\ref{arch-repre}). Then the splatted portrait foreground features are seamlessly integrated with the background features using our Neural Rerenderer to finish an in-the-wild head swap result (Sec.~\ref{BFMB}).

% To ensure high identity preservation, we first describe a series of finetuning strategies to adapt the 2D portrait generation model to the source head domain, generate a batch of training data through inpainting and re-extract the SMPL-X parameters from the training data(Sec.~\ref{arch-gen}).  To ensure consistency across frames, we propose an in-the-wild 4D portrait representation utilizing 3DGS and holistic human body priors(Sec.~\ref{arch-repre}). For consistency high quality rendering, we introduce temporal conditional feature on the Neural Gaussian Texture mechanism(Sec.~\ref{TCF}). To fill the mismatch region naturally, we proposed a background feature map blending method to get the in-the-wild head swap result(Sec.~\ref{BFMB}).

\subsection{Preliminary}
\label{sec:preliminary}
\subsubsection{3D Gaussian Splatting}

3DGS chooses 3D Gaussians as geometric primitives to represent scenes. Every Gaussian is defined by a 3D covariance matrix $\mathbf{\Sigma}$ centered at point $\mathbf{x_0}$:
\begin{equation}
  g(\mathbf{x}) = e^{-\frac{1}{2} (\mathbf{x}-\mathbf{{x_0}})^{T} \mathbf{\Sigma} ^{-1}(\mathbf{x}-\mathbf{x_0})}.
  \label{eq:gs}
\end{equation}
$\mathbf{\Sigma}$ is decomposed into a rotation matrix $R$ and a scaling matrix $\varLambda$ corresponding to learnable quaternion $\mathbf{q}$ and scaling vector $\mathbf{s}$:
\begin{equation}
  \mathbf{\Sigma} = R\varLambda\varLambda^TR^T.
  \label{eq:cov}
\end{equation}

Each 3D Gaussian is attached another two attributes: opacity $o$ and SH coefficients $\mathbf{h}$. The final color for a given pixel is calculated by sorting and blending the overlapped Gaussians:
\begin{equation}
  \mathbf{C} = \sum_{i\in N} \mathbf{c}_i\alpha_i\prod_{j=1}^{i-1} (1-\alpha_j),
  \label{eq:blend}
\end{equation}
where $\alpha_i$ is computed by the multiplication of projected Gaussian and $o$. Gaussian field can be denoted as  $\{\mathbf{x_0}, \mathbf{q}, \mathbf{s}, o, \mathbf{h}\}$.

\subsubsection{SMPL-X}

SMPL-X model~\cite{SMPL-X:2019} is a holistic, expressive body model, and is defined by a function $M\left(\beta, \theta, \psi \right):\mathbb{R}^{| \beta | \times | \theta | \times | \psi |} \rightarrow \mathbb{R}^{3V}$: 
\begin{align}
M   \left(\beta,\theta,\psi\right) &= W \left(T \left(\beta,\theta,\psi \right) , J\left(\beta \right),  \theta,\mathcal{W} \right), 									\label{skinning}			\\
T \left(\beta,\theta,\psi\right) &= \bar{T} + B_S \left(\beta;\mathcal{S} \right) + B_E  \left( \psi;\mathcal{E} \right) + B_P \left(\theta;\mathcal{P} \right).	\label{smpl_offsets}
\end{align}
  $\beta$, $\theta$, $\psi$ are shape, pose and expression parameters, respectively. $B_S \left(\beta;\mathcal{S} \right)$, $ B_P \left(\theta;\mathcal{P} \right)$, $ B_E \left(\psi;\mathcal{E} \right)$ are the blend shape functions. Blend skinning function $W(\cdot)$~\cite{lewis2000pose} rotates the vertices in $T\left(\cdot\right)$ around the estimated joints $J(\beta)$ smoothed by blend weights. To model long hairs and loose clothing, we introduce a learnable vertices displacement and the final mesh is computed as:
\begin{equation}
    \widehat{M}   \left(\beta,\theta,\psi\right) = M   \left(\beta,\theta,\psi\right) + \Delta M.
\end{equation}

\subsection{Image Head Swap Dataset Generation}\label{arch-gen}
\begin{figure*}[t]
  \centering
  \includegraphics[width=\linewidth]{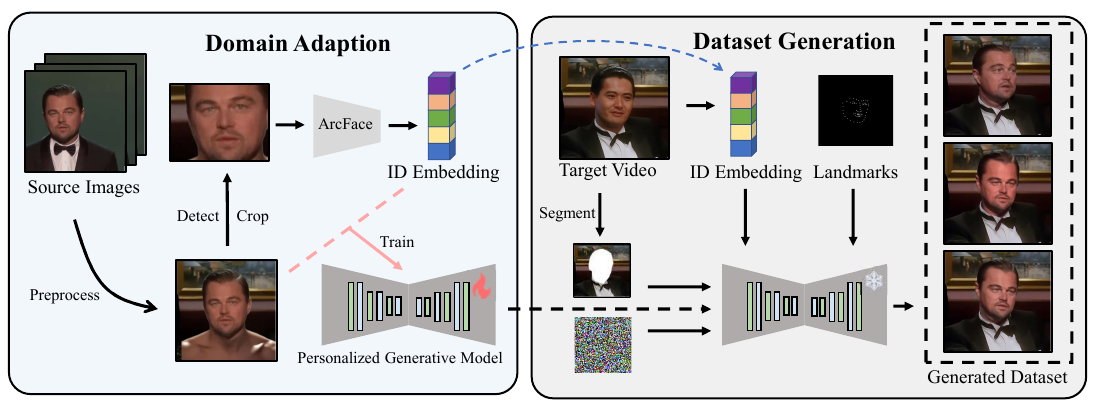}
  \caption{We begin by performing domain adaptation on a pre-trained 2D head generation model. Subsequently, we use this model to inpaint the head region in the target video, thus creating a dataset suitable for head swapping applications. We will mosaic some pictures when publishing.}
  \label{fig:sec3.1}
  \vspace{-2mm}
\end{figure*}

As illustrated in Fig.~\ref{fig:sec3.1}, we utilize a 2D head generation model to inpaint the head region of the target video, thereby constructing a dataset for head swapping. Although existing human head generation models have demonstrated proficiency in tasks that require identity-consistent head generation, they often fall short in preserving the detailed and personalized attributes of the source subject's head.  To address this limitation, we finetune a pretrained head generation model Arc2Face~\cite{paraperas2024arc2face} with few-shot source input images $\mathcal{I}_{\textrm{src}}$ to enhance identity similarity following ~\cite{ruiz2023dreambooth}.

% Given source images $I_{\textrm{src}}$, we first preprocess them to extract personal identity information to the greatest extent possible. Then We train
% a personalized DreamBooth model $\hat{\mathcal{D}}$ on the processed head images and inpaint on $I_{T}$ to generate head swap datasets. Moreover, we re-extract the SMPLX parameters of the generated dataset.

Before finetuning, we first need to isolate background and clothing information from portrait images $\mathcal{I}_{\textrm{src}}$. We achieve this by cropping the foreground human region from the original images and overlaying it onto the background of the target video. We employ a diffusion model to inpaint it to remove the clothes. These strategies are important in eliminating the potential leakage of unrelated information, such as background and clothing details, ensuring that the fine-tuning process focuses solely on the relevant facial and head regions. We follow Dreambooth~\cite{ruiz2023dreambooth} to finetune the portrait generation model and use ControlNet~\cite{zhang2023adding} to inject landmarks to control head poses and expressions. Then we inpaint the head region of $\mathcal{I}_{\textrm{tgt}}$ with the finetuned model to get the head swap dataset $\mathcal{I}_{\textrm{gen}}$. The mask of the head region is warped from the head mask of coarse reenactment result, according to the 5 face key points. Please refer to the supplementary material for more details.

\begin{figure*}[t]
  \centering
  \includegraphics[width=\linewidth]{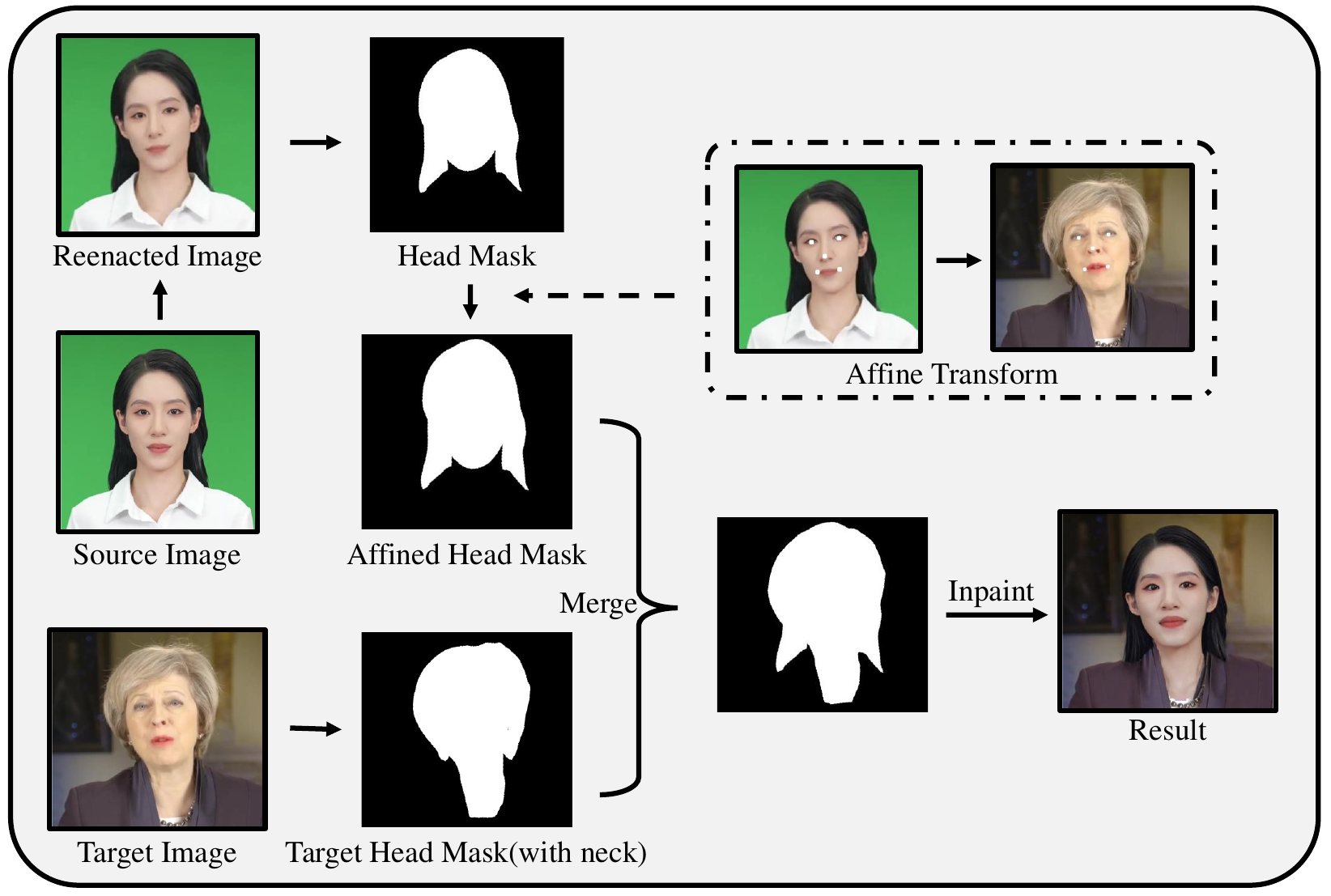}
  \caption{We first employ a coarse face reenactment method to obtain a head with the corresponding pose and expression. Then, we warp the head mask of the face reenactment result based on the five key points between the reenacted head and the target head.}
  \label{fig:warp}
  \vspace{-2mm}
\end{figure*}

In the inpainting process, the head mask region should not only cover the head mask of the target image but also accommodate the shape of the source head. As illustrated in Fig.~\ref{fig:warp}, we first employ a coarse face reenactment method\cite{hong23implicit} to obtain a head with the corresponding pose and expression. Then, we warp the head mask of the face reenactment result based on the five key points between the reenacted head and the target head.

\subsection{Dynamic Portrait Representation}
Due to the stochastic nature of the diffusion model, the generated head swap frames often exhibit inconsistencies.
Following previous 3D Gaussian portrait models \cite{xiang2024flashavatar,Gao2024PortraitGen,qian2023gaussianavatars}. We distill the knowledge of the adapted diffusion model in Sec.~\ref{arch-gen} into a 3D representation to enhance the consistency. Firstly we maintain a 3D Gaussian field on the UV space of a SMPL-X~\cite{SMPL-X:2019} surface $M   \left(\beta,\theta,\psi\right)$ with learnable vertices offset $\Delta M$, and further deform the Gaussians according to the deformation of underlying meshes tracked from the input video. By embedding a 3D Gaussian field on the surface, the 3D Gaussian field could be efficiently transformed by SMPL-X shape, pose, and expression parameters $\beta$, $\theta$, $\psi$. Then we store learnable features for each Gaussian. To be specific, we have a Neural Gaussian Field $\mathcal{G}$ in the UV field where each pixel is characterized by four attributes: neural feature, opacity, scales, and rotation. Using UV mapping $T_{\textrm{uv}} \in \mathbb{R}^3 \rightarrow \mathbb{R}^2$, we transform neural Gaussians from UV space to 3D space.

This operation $\mathcal{F}$ could be written as:
\begin{equation}
    (X_0, Q, S, O, F) = \mathcal{F}(M   \left(\beta,\theta,\psi\right) + \Delta M , T_{\textrm{uv}}, \mathcal{G}),
\end{equation}
Given $M   \left(\beta,\theta,\psi\right)$, $T_{\textrm{uv}}$ and $\mathcal{G}$, we could get the embedded 3D Gaussian field $(X_0, Q, S, O, F)$ corresponding to the point position, rotation, scale, opacity, and neural feature field of a certain frame.

With camera intrinsic parameters $K$, camera poses $P = \{P_i\}_{i=1}^{T}$, and the 3D Gaussian field, we perform differentiable tile renderer $\mathcal{R}$ to render a foreground portrait feature image $\phi_{\textrm{fore}}$ and alpha map $A_{\textrm{out}}$:

\begin{align}
    \phi_{\textrm{fore}} ,A_{\textrm{out}} = \mathcal{R}(& (X_0, Q, S, O, F), K, P),
\end{align}

\subsubsection{Handling Inconsistencies}\label{arch-repre}

Although we employ ControlNet to inject landmarks into the generation process, there still exist some expression and pose misalignments between $\mathcal{I}_{\textrm{gen}}$ and $\mathcal{I}_{\textrm{tgt}}$.
To mitigate these issues, we retrack the SMPL-X parameters ($\beta_{\textrm{gen}}$, $\theta_{\textrm{gen}}$, $\psi_{\textrm{gen}}$) of $\mathcal{I}_{\textrm{gen}}$ during the training phase so that the training data of our portrait representation model is accurate. The parameters tracked from the original target video $\mathcal{I}_{\textrm{tgt}}$ are then used for inference.

\begin{figure*}[t]
  \centering
  \includegraphics[width=0.95\linewidth]{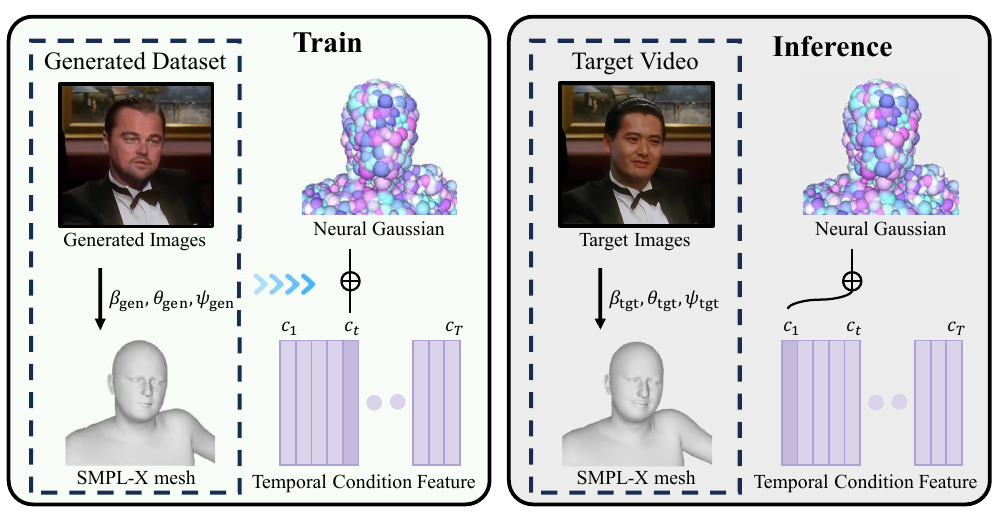}
  \caption{To mitigate the expression and pose misalignments, we retrack the SMPL-X parameters ($\beta_{\textrm{gen}}$, $\theta_{\textrm{gen}}$, $\psi_{\textrm{gen}}$) of $\mathcal{I}_{\textrm{gen}}$ during the training phase. The parameters tracked from the original target video $\mathcal{I}_{\textrm{tgt}}$ are then used for inference.}
  \label{fig:sec3.2}
  \vspace{-2mm}
\end{figure*}

Further, to resolve inconsistencies across different frames in the generated dataset, we draw inspiration from recent advancements in dynamic scene modeling~\cite{park2021hypernerf}. As illustrated in Fig.~\ref{fig:sec3.2}, during the training stage, we store a set of learnable features, $C = \{c_{t}|t \in [1, T]\}$, which are broadcasted to every Gaussian across all frames. For inference, the feature $c_{1}$ is utilized consistently across all frames, enhancing the temporal coherence and stability of the generated sequences.

% Given camera intrinsic parameters $K$, camera poses $P = \{P_i\}_{i=1}^{T}$, and the 3D Gaussian field, we perform differentiable tile renderer $\mathcal{R}$ to render a foreground portrait feature image $\phi_{\textrm{fore}}$ and alpha map $A_{\textrm{out}}$:
% \begin{align}
%     (X_0, Q, S, O, F) = &\mathcal{F}(M   \left(\beta_{\textrm{gen}},\theta_{\textrm{gen}},\psi_{\textrm{gen}}\right) 
%     \\&+ \Delta M , T_{\textrm{uv}}, \mathcal{G}),\notag
%     \\
%     \phi_{\textrm{fore}} ,A_{\textrm{out}} = \mathcal{R}(& (X_0, Q, S, O, F, c_{t}), K, P),
% \end{align}
\subsubsection{Neural Rerendering}\label{BFMB}

Our portrait representation effectively models the swapped head, yet reintegrating this 3D head into the original video remains challenging. Notably, significant mismatches often occur in the head region between $\phi_{\textrm{fore}}$ and $I_{\textrm{tgt}}$ due to differences in head shapes and hairstyles. So we designed a neural rerendering module to handle this mismatch.

We first remove the foreground head region in $I_{\textrm{tgt}}$ and inpaint it with Inpaint-Anything~\cite{yu2023inpaint}. Then the inpainted image $\hat{I}_{\textrm{tgt}}$ (with only background and torso) is operated by a 2D background encoder $\mathcal{E}_{\textrm{bg}}$ to convert it to the feature domain. Based on the overlapping area between the rendered alpha channel $A_{\textrm{out}}$ and the head mask $M_{\textrm{head}}$, we fuse the foreground head and  background on the feature domain:
\begin{align}
    M_{\textrm{fore}} &= A_{\textrm{out}} \cap M_{\textrm{head}}, \\
    \phi_{\textrm{bg}} &= \mathcal{E}_{\textrm{bg}}(\hat{I}_{\textrm{tgt}}), \\
    \phi_{\textrm{fused}} &= M_{\textrm{fore}} * \phi_{\textrm{fore}} + (1 - M_{\textrm{fore}}) * \phi_{\textrm{bg}},
\end{align}

Then the fused feature map is operated by a 2D Neural Rerenderer $\mathcal{U}$ to convert it to RGB domain:
\begin{align}
    I_{\textrm{out}} &= \mathcal{U}(\phi_{\textrm{fused}}).
\end{align}

$M_{\textrm{fore}}$ is the overlapping area between the rendered alpha channel and the head mask.$\phi_{\textrm{fore}}$, $\phi_{\textrm{bg}}$ and $\phi_{\textrm{fused}}$ share the same resolution.

\subsubsection{Training Strategy}
We train the head swap dynamic Gaussian representation with the following loss terms:

\paragraph{Reconstruction Loss} This loss requires that the rendered result is consistent with the input RGB image, which is common for RGB reconstruction and can be formulated as:
\begin{equation}
L_{\textrm{recon}}(I_{\textrm{out}},I_{\textrm{gen}}) =  \|I_{\textrm{out}} - I_{\textrm{gen}} \|_1.
\end{equation}

% This part is divided into two stages. In the initial stage of training, $I$ is rendered from $\beta_{\textrm{gen}}$, $\theta_{\textrm{gen}}$, $\psi_{\textrm{gen}}$. After training a coarse Gaussian field from the initial stage, the coarse head swap images can be rendered at each frame, then we add small noise on these images, and after the diffusion progress of $\hat{\mathcal{D}}$ to update training datasets and replace $I_{\textrm{gen}}$ with them. At this stage, $I$ is rendered from $\beta_{\textrm{tgt}}$, $\theta_{\textrm{tgt}}$, $\psi_{\textrm{tgt}}$, which is the SMPL-X parameters extract from $\mathcal{I}_{\textrm{tgt}}$.

\paragraph{Mask Loss} This loss requires that the rendered alpha channel $A$ is consistent with the segmentation map of the input source image:
\begin{equation}
L_{\textrm{mask}}(A_{\textrm{out}},A_{\textrm{gen}}) =  \|A_{\textrm{out}} - A_{\textrm{gen}} \|_1.
\end{equation}

\paragraph{Perceptual Loss}
The perceptual loss $L_{\textrm{LPIPS}}$ of \cite{zhang2018perceptual} is utilized to provide robustness to slight misalignments and shading variations and improve details.
We choose VGG as the backbone of LPIPS.

\paragraph{Background Loss}
The background loss is utilized to prevent the color shift of background arised from the neural rerendering module.
\begin{equation}
L_{\textrm{back}}(I_{\textrm{out}},I_{\textrm{tgt}}) =  \|I_{\textrm{out}} * (1 - M_{\textrm{fore}}) - I_{\textrm{tgt}} * (1 - M_{\textrm{fore}}) \|_1.
\end{equation}

In summary, the overall loss of training in our model is defined as:
\begin{equation}
\begin{split}
  L_{\textrm{total}} =& \lambda_{1}L_{\textrm{recon}}(I,I_{\textrm{gen}})+\lambda_{2}L_{\textrm{mask}}(A,A_{\textrm{gen}}) \\&+ \lambda_{3}L_{\textrm{LPIPS}}(I,I_{\textrm{gen}}) + \lambda_{4}L_{\textrm{back}}(I_{\textrm{out}},I_{\textrm{tgt}})
\end{split}
\end{equation}

We add a loss item guided by the super-resolution module in FaceChain\cite{liu2023facechain} for higher resolution after training for 10 epochs.
\section{Experiments}

% \includegraphics[page=1,width=\linewidth]{figure/swap4d_comp1_crop.pdf}

% \begin{figure}[t]
%   \centering
%   \includegraphics[width=\linewidth]{figure/sec3.1_v1.pdf}
%   \caption{.}
%   \label{fig:sec3.1}
%   \vspace{-2mm}
% \end{figure}

\begin{figure*}[t]
  \centering
  \includegraphics[page=1,width=\textwidth]{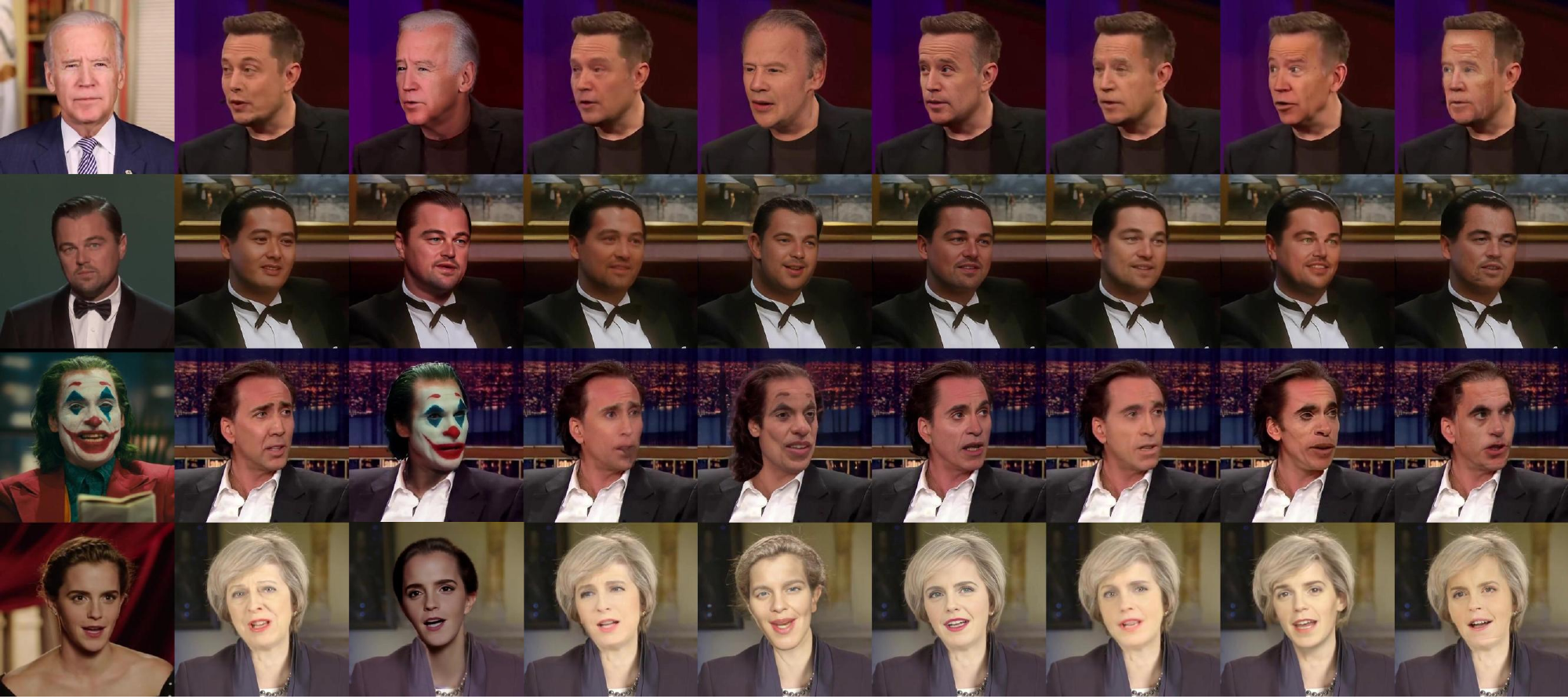}
  \noindent
  \raisebox{1mm}{
    \makebox[\dimexpr\textwidth/9-0.7em][c]{Source} \hspace{0pt}
    \makebox[\dimexpr\textwidth/9-0.7em][c]{Target} \hspace{0pt}
    \makebox[\dimexpr\textwidth/9-0.7em][c]{Ours} \hspace{0pt}
    \makebox[\dimexpr\textwidth/9-0.7em][c]{DiffSwap} \hspace{0pt}
    \makebox[\dimexpr\textwidth/9-0.7em][c]{HeSer} \hspace{0pt}
    \makebox[\dimexpr\textwidth/9-0.7em][c]{DeepLiveCam} \hspace{0pt}
    \makebox[\dimexpr\textwidth/9-0.7em][c]{BlendFace} \hspace{0pt}
    \makebox[\dimexpr\textwidth/9-0.7em][c]{FaceAdapter} \hspace{0pt}
    \makebox[\dimexpr\textwidth/9-0.7em][c]{InfoSwap} \hspace{0pt}
  }
  \vspace{-5mm}
  \caption{Our method leverages a dynamic Gaussian portrait framework integrated with adapted diffusion portrait priors, achieving better quality and remarkable consistency.}
  \label{fig:comp1}
  % \vspace{2mm}
\end{figure*}

\subsection{Implementation Details}
We change the input channels of the official Gaussian splatting code into 32 and use it to get the splatted feature map. Our neural rerenderer is implemented as a 3-layer CNN and the layer-wise channel configuration is (8,8,8,3), we use the ReLU activation function after each convolution, and we set kernel size as 3 and padding number as 1 for all convolution layers. Our background neural encoder is implemented as a 2-layer CNN and the layer-wise channel configuration is (8,8,5), then we concat the input background with the output of CNN to get the background feature map. The channels of temporal condition feature is also 8.

We collected some videos and source images from the Internet and captured some source images by ourselves. We use SDXL\cite{rombach2022highresolution} to undress the source images with the prompt ``nude, no clothes, a person without clothes''. We use FaRL\cite{Zheng_2022_CVPR} to get head masks. To get background images of target frames, we use Inpaint-Anything~\cite{yu2023inpaint} to inpaint the segmented human region. We use an algorithm similar to TalkSHOW\cite{yi2022generating} for fitting SMPL-X parameters to video frames. Our experiments are run on one RTX 4090 GPU. $\lambda_1, \lambda_2, \lambda_3$ are set to  1.0, 1.0, 1.0.

\begin{table}[h] %[h] %[!htb] %[h!]
  \caption{Inference Speed. (Liveportrait here is the method of ``single-image head swap + LivePortrait'')}
  \label{tab:IS}
  \centering
   \renewcommand{\arraystretch}{1.2}
   \renewcommand{\tabcolsep}{3pt} % {1.05mm} %{1.5mm} % (1.5mm = 4.252 pts)
  \resizebox{1.0\linewidth}{!}{
  \footnotesize
  {
      \begin{tabular}{@{}lccccccccc@{}}
        \toprule
         Method &  DiffSwap & DeepLiveCam & BlendFace & InfoSwap & FaceAdapter & Heser & LivePortrait$^{*}$ & Ours  \\ \midrule
         FPS $\uparrow$ & 0.2 & 60 & 0.9 & 1.0 & 0.3 & 0.1 & 30 & 76 \\
        \bottomrule
      \end{tabular}}
  }
\end{table}

All experiments were performed on a single NVIDIA RTX 4090 GPU. To quantify the computational requirements, we measured the processing time and GPU memory consumption for each stage when handling a target video sequence of 500 frames. The detailed resource utilization is as follows: (1) Arc2Face fine-tuning requires approximately 5 minutes with 15GB GPU memory allocation; (2) Dataset generation requires approximately 4 minutes with 15GB GPU memory allocation; (3) SMPL-X tracking requires approximately 2 minutes with 5GB GPU memory allocation; and (4) The fitting procedure requires approximately 8 minutes with 10GB GPU memory allocation. Most compared methods are feed-forward systems, whereas GSwap and the ``Single-Image Head Swap + LivePortrait'' method require a fine-tuning process. However, in the inference stage, our method is significantly faster than these feed-forward approaches. For a comprehensive comparison of computational efficiency, Table~\ref{tab:IS} presents the rendering speeds of all methods on an RTX 4090 GPU, which quantitatively demonstrates the advantage of our method in inference efficiency.

Regarding the training cost, only the pretrained 2D portrait generative model requires fine-tuning. Auxiliary tools (e.g., FaRL for head masking, Inpaint-Anything for background inpainting, ControlNet for landmark injection) are lightweight, off-the-shelf utilities that do not require training. They act as data preprocessors to ensure high-quality input for the domain-adapted diffusion model, rather than as independent components of the core pipeline. The entire data preprocessing stage completes in less than 5 minutes.

\subsection{Headswap Result}
We compare our proposed GSwap with the state-of-the-art head swapping model Heser\cite{shu2022head} and face swapping models DeepLiveCam\cite{deeplivecam}, DiffSwap\cite{zhao2023diffswap}, InfoSwap~\cite{Gao_2021_CVPR}, BlendFace~\cite{Shiohara_2023_ICCV}, FaceAdapter~\cite{han2024face}, LivePortrait~\cite{guo2024liveportrait}. HeSer is not open source. We emailed the authors and they suggest using an unofficial version. For more vivid head swapping results generated by our method please refer to the supplementary video. LivePortrait primarily focuses on the portrait animation task rather than the head-swapping task. However, to further demonstrate the superiority of our approach, we adapted LivePortrait for this context by conducting a comparative experiment using a combined pipeline of ``Single-Image Head Swap + LivePortrait.'' The single-image head swap method utilized in this baseline is described in Sec.~\ref{arch-gen}.

\begin{figure}[t]
  \centering
  \includegraphics[width=\linewidth]{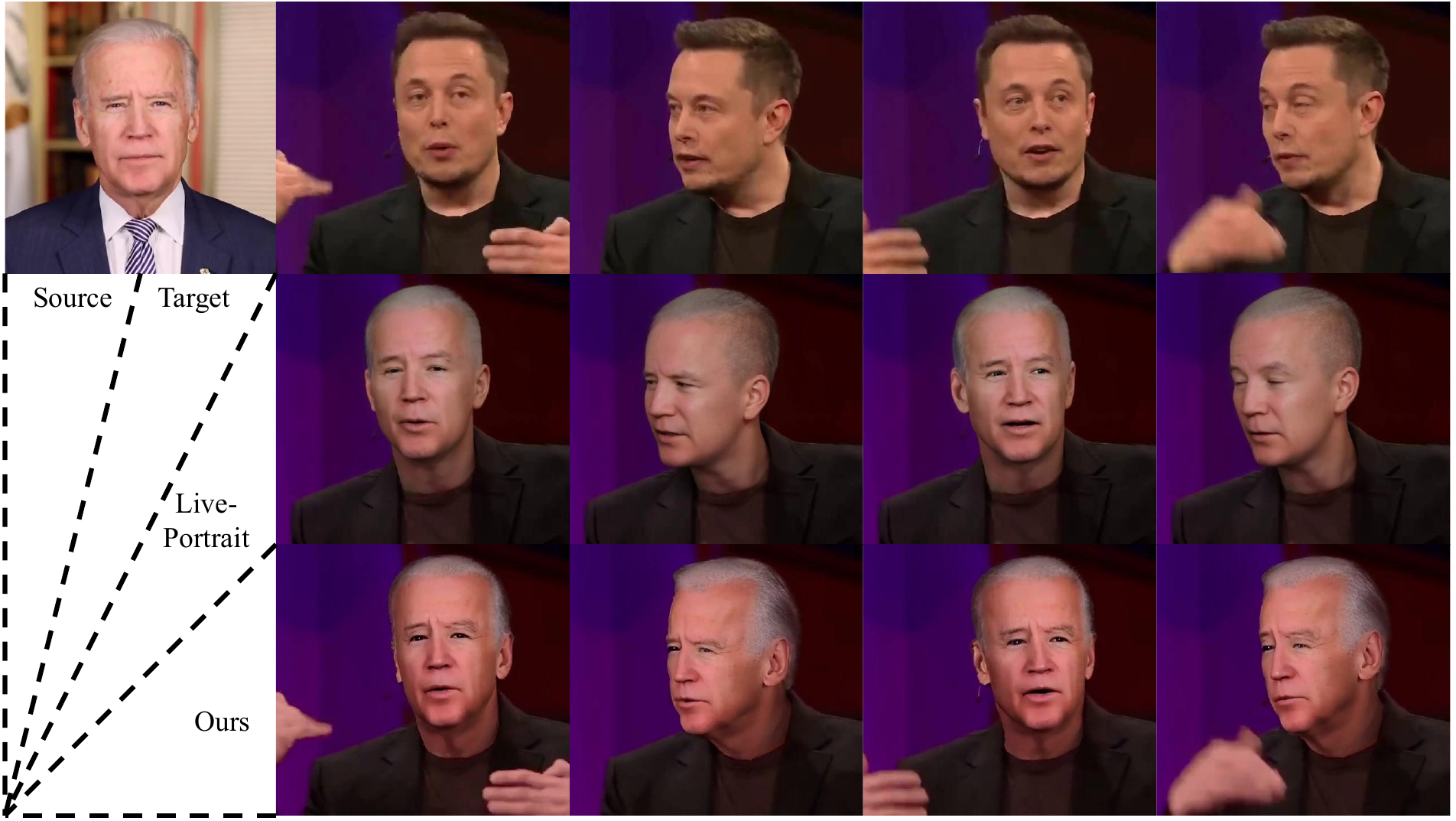}
  \caption{Qualitive Comparison with the combined method of ``Single-Image Head Swap + LivePortrait''.}
  \label{fig:lp_comp}
  \vspace{-2mm}
\end{figure}

\noindent\textbf{Qualitative Comparison.} 
The results shown in Fig.\ref{fig:comp1} highlight the advantages of our method over existing techniques in various dimensions such as identity preservation, pose and expression consistency, skin color alignment, head-background coherence, and overall fidelity. For both Heser and our head swapping model, the source images consist of the same 4-12 frames. DeepLiveCam and DiffSwap do not support the use of multiple reference images. Therefore, we select a single frame from the few-shot images to serve as the reference for these methods. We note that both DeepLiveCam and Heser typically introduce only minor modifications to the target subject, potentially due to unstable training and the limited capabilities of the GAN model. BlendFace and InfoSwap can hardly change face shapes. Moreover, DiffSwap and FaceAdapter struggles with poor temporal coherence, a problem arising from the stochastic nature of diffusion models. The combined method of ``Single-Image Head Swap + LivePortrait'' fails to maintain consistency in surrounding regions (such as hands and shoulders) between the generated results and the target video. Furthermore, it struggles to preserve identity consistency. Our model leverages a dynamic Gaussian portrait framework integrated with adapted diffusion portrait priors, achieving better quality and remarkable consistency.

% The results in Fig. \ref{fig:comp1} underscore the superiority of our approach over existing methods across multiple aspects, e.g., identity preservation, pose and expression consistency, skin color alignment, head-background coherence, and fidelity. 
% Source images for Heser and our head swapping model are the same 4-12 frames. The source image for face swapping methods DeepLiveCam and DiffSwap is one of the few-shot images. We observe that both DeepLiveCam and Heser could only introduce minor changes to target subject, this may be caused by the unstable training and limited capacity of GAN model. Furthermore, we found DiffSwap suffer from bad temporal coherence, which originates from the stochastic nature of diffusion models. Our model distills powerful pretrained diffusion portrait priors to 4D Gaussian portrait model, achieving both high quality and remarkable consistency.

% The face swapping methods can swap facial elements from source image to target video, the face shape can be slightly transferred; however, they can not handle the large differences between face shapes. The skin color and hair style also restrict the identity similarity in face cognition.
% Heser tends to perform poorly when the source images is largely different from the target video. Restricted by the limited representation ability of 2D GAN models, its identity similarity is inferior to ours. We also perform better temporal consistency than Heser because of our 4D representation.

\begin{table}[t] %[h] %[!htb] %[h!]
  \caption{Quantitive Results.}
  \label{tab:csimapdaed}
  \centering
   \renewcommand{\arraystretch}{1.2}
   \renewcommand{\tabcolsep}{3pt} % {1.05mm} %{1.5mm} % (1.5mm = 4.252 pts)
  \resizebox{1.0\linewidth}{!}{
  \footnotesize
      \begin{tabular}{@{}lccccc@{}}
        \toprule
         Method & CSIM $\uparrow$ & AED $\downarrow$ & APD $\downarrow$ & CLIP Score $\uparrow$ \\ \midrule
         InfoSwap~\cite{Gao_2021_CVPR}  & 0.494 & 0.314 & 0.008 & 99.30\% \\
         BlendFace~\cite{Shiohara_2023_ICCV} & 0.501  & 0.249 & 0.006 & 99.46\% \\
         FaceAdapter~\cite{han2024face} & 0.533   & 0.276 & 0.008 & 99.34\% \\
         DiffSwap~\cite{zhao2023diffswap}    & 0.258 & \textbf{0.235} & 0.009 & 95.20\% \\
         DeepLiveCam~\cite{deeplivecam}    & 0.712 & 0.251 & \textbf{0.005} & 99.27\% \\ \midrule
         LivePortrait~\cite{guo2024liveportrait} & 0.381 & 0.281 & 0.006 & 99.60\% \\ \midrule
         Heser~\cite{shu2022head} & 0.280  &  0.343 & 0.017 & 98.58\% \\
         Ours     & \textbf{0.727} & 0.286  & 0.009 & \textbf{99.64\%} \\
        \bottomrule
      \end{tabular}
  }
\end{table}

\begin{table}[t] %[h] %[!htb] %[h!]
  \caption{Average score from the user study, rating from 0 to 3.}
  \label{tab:avg_score}
  \centering
   \renewcommand{\arraystretch}{1.2}
   \renewcommand{\tabcolsep}{3pt} % {1.05mm} %{1.5mm} % (1.5mm = 4.252 pts)
  \resizebox{1.0\linewidth}{!}{
  \footnotesize
      \begin{tabular}{@{}lcccccc@{}}
        \toprule
         Method & ID $\uparrow$ & {Exp} $\uparrow$ & Skin Color $\uparrow$ & Inpainting $\uparrow$ & Holistic $\uparrow$  \\ \midrule
         DiffSwap~\cite{zhao2023diffswap}    & 0.40          & 0.72 & 0.49 & -- & 0.40 \\
         DeepLiveCam~\cite{deeplivecam}    & 1.80          & \textbf{2.55} & 1.67 & -- & 2.17 \\
         Heser~\cite{shu2022head} & 1.08          &  0.70         & 1.25 &  0.60      & 0.85 \\
         Ours                                    & \textbf{2.72} & 2.03         & \textbf{2.59} & \textbf{2.40} & \textbf{2.57} \\
        \bottomrule
      \end{tabular}
  }
\end{table}

\noindent\textbf{Quantitative Comparison.} 
The quantitative results are shown in Table.~\ref{tab:csimapdaed}. We collect 30 pairs of source and target from the Internet as the test dataset. Three metrics are included: 

(1) \textbf{CSIM:} CSIM measures the identity preservation between two images, through the cosine similarity of two embeddings from a pretrained face recognition network~\cite{deng2019arcface}. 

(2) \textbf{AED:} AED is the mean $\mathcal{L}_1$ distance of the expression parameters between the animated and the driving images. These parameters, which include facial movement, eyelid, and jaw pose parameters, are extracted by the state-of-the-art 3D face reconstruction method SMIRK~\cite{retsinas20243d}. 

(3) \textbf{APD:} APD is the mean $\mathcal{L}_1$ distance of the pose parameters between the animated and the driving images. The pose parameters are extracted by SMIRK~\cite{retsinas20243d}. 

(4) \textbf{CLIP Score (Frame Consistency):} The CLIP Score (Frame Consistency)~\cite{zhang2023towards} measures the temporal continuity between consecutive frames in a video by computing the cosine similarity of image embeddings from a pre-trained CLIP model~\cite{radford2021learning}. These embeddings, which effectively capture high-level semantic and visual features, are used to quantify the similarity between adjacent frames, thereby evaluating the temporal consistency of the generated video sequences.

InfoSwap, BlendFace, FaceAdapter, DiffSwap and DeepLiveCam are face swapping methods, our method and Heser are for head swapping. Face swapping methods directly injects facial components into the target image, achieving superior expression similarity. We address a significantly more challenging head swapping task, achieving superior performance on CSIM metrics while minimizing the impact on facial expressions and poses compared to existing methods.

We conduct a user study for quantitative comparison of our GSwap with the face swapping methods. Considering the potential cognitive overload that may arise from evaluating an excessive number of methods, we strategically selected three representative comparison methods based on the quantitative metrics presented in Table.~\ref{tab:csimapdaed}.  We ask the users to rank
\begin{itemize}
\item \textbf{ID}: The identity similarity with the source images.
\item \textbf{Exp}: The emotion and pose similarity with the target images.
\item \textbf{Skin Color}: How well the skin color is similar to source images in each method.
\item \textbf{Inpainting}: The inpainting smoothness between the generated head and background.
\item \textbf{Holistic}: The holistic quality of the generated frames considering the above four aspects.
\end{itemize}

We collected statistics from 50 participants across 10 groups of head/face swap results. For each case, the video results were randomly shuffled for fair comparison. The participants rank the methods according to the questions, and the rankings are scaled into scores of zero to three. Different from Heser\cite{shu2022head}, we consider skin color as an identity information, thus it should be similar to the source images.
As shown in Table.~\ref{tab:avg_score}, our method outperforms other methods by a large margin in identity similarity, skin color similarity, inpainting smoothness, and holistic quality.

\begin{figure}[htb]
  \centering
  \setlength{\abovecaptionskip}{0.cm}
  \includegraphics[width=0.9\linewidth]{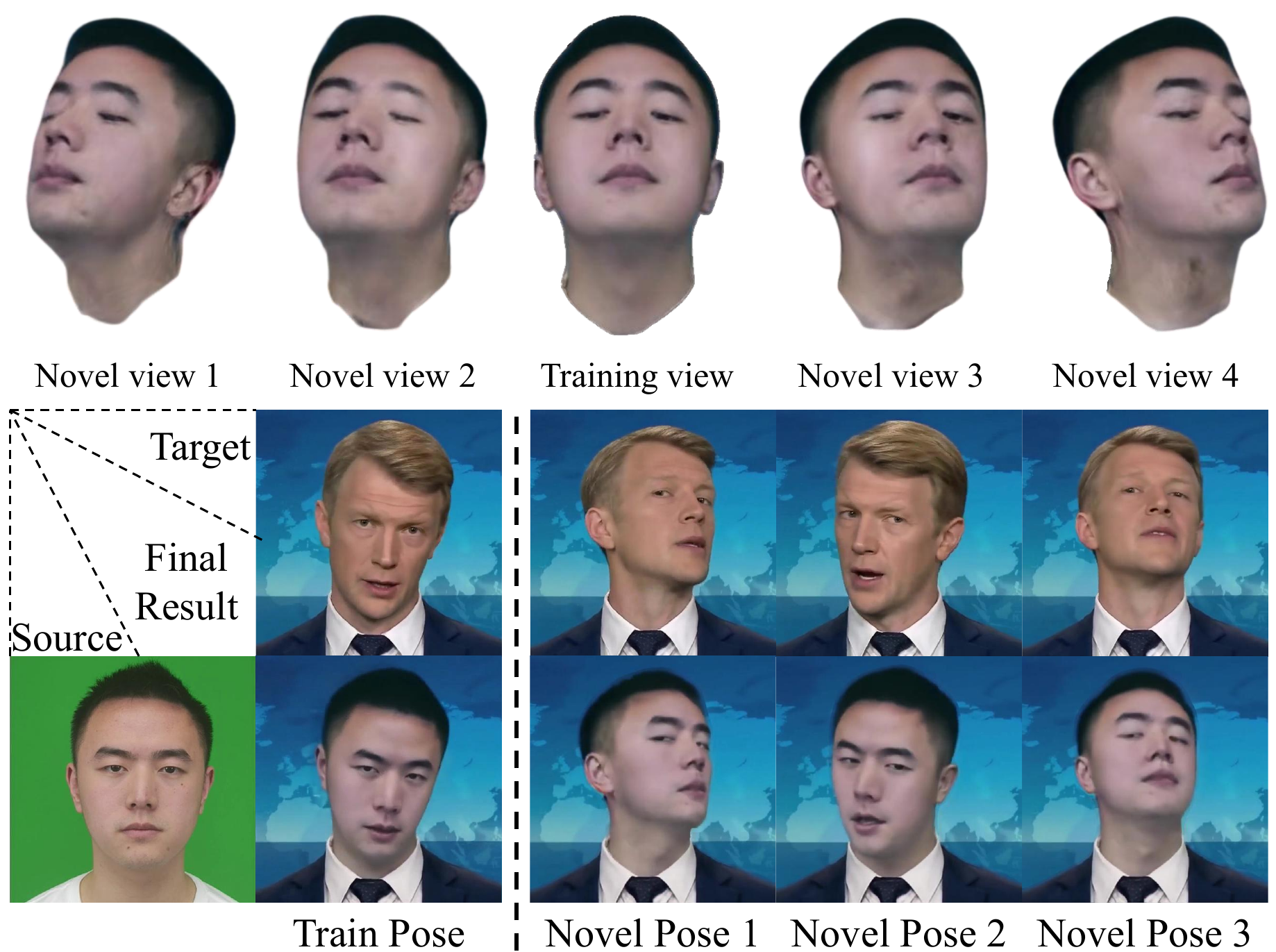}
  \caption{To validate our 3D consistency, we provide rendered results under both novel view and novel pose conditions. Since the novel view synthesis lacks corresponding background input, only the head region is displayed in the multi-view results.}
  \label{fig:nvnp}
\end{figure}

\begin{figure}[htb]
  \centering
  \setlength{\abovecaptionskip}{0.cm}
  \includegraphics[width=0.95\linewidth]{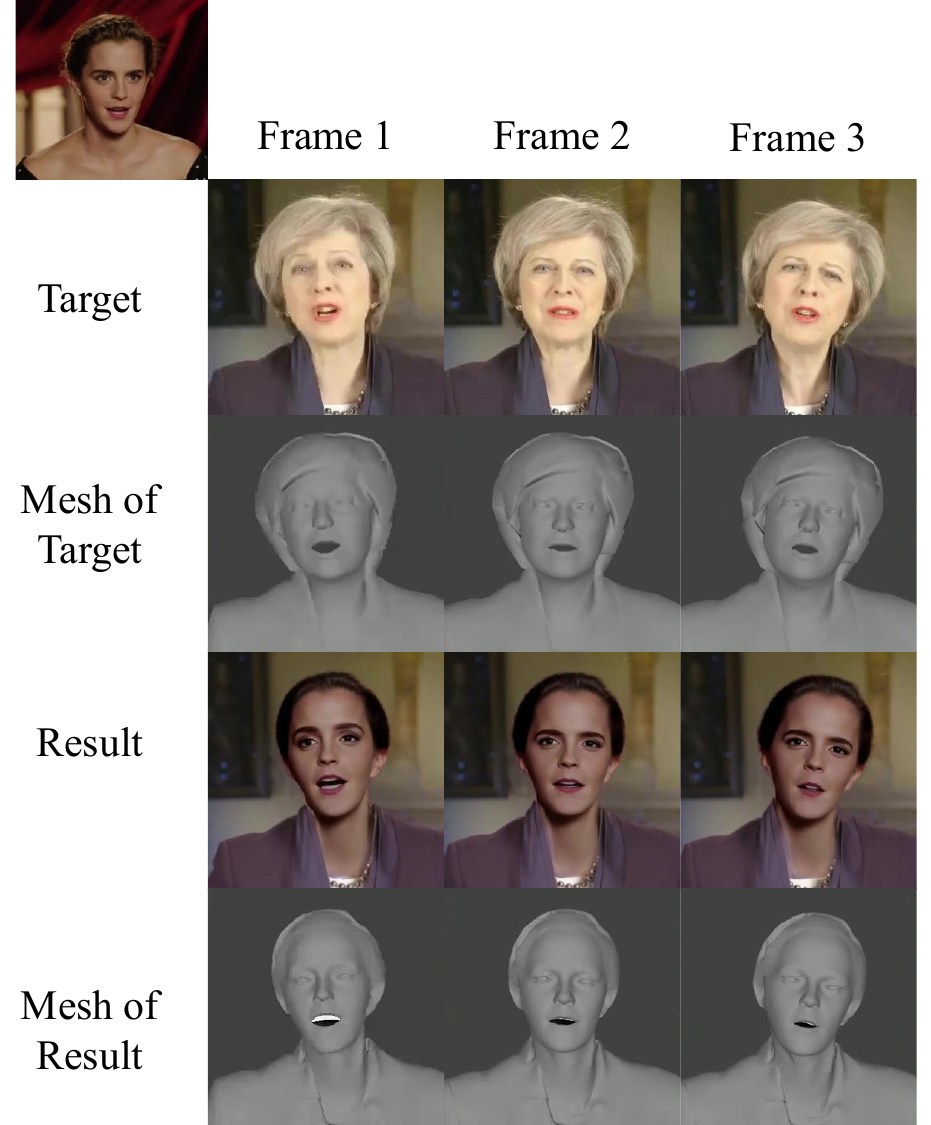}
  \caption{The deformed mesh aligns with the geometry of the real human face, effectively mitigating potential error accumulation caused by the discrepancy between the SMPL-X facial representation and the ground-truth geometry.}
  \label{fig:mesh}
  \vspace{-2mm}
\end{figure}

\noindent\textbf{3D Rendering Results.} Fig.~\ref{fig:nvnp} shows multi-view and novel-pose results. As the novel view lacks background input, only the head part is displayed in the multi-view results. We bind 3D Gaussians to the mesh during the distillation process and explicitly deform the mesh during optimization. As illustrated in Fig.~\ref{fig:mesh}, the deformed mesh aligns consistently with the ground-truth facial geometry. This alignment effectively mitigates potential error accumulation caused by discrepancies between the parametric SMPL-X representation and the real human face.

\subsection{Ablation Study}
\subsubsection{Handle Inconsistency of Generated Dataset}
\begin{figure}[t]
  \centering
  \includegraphics[width=\linewidth]{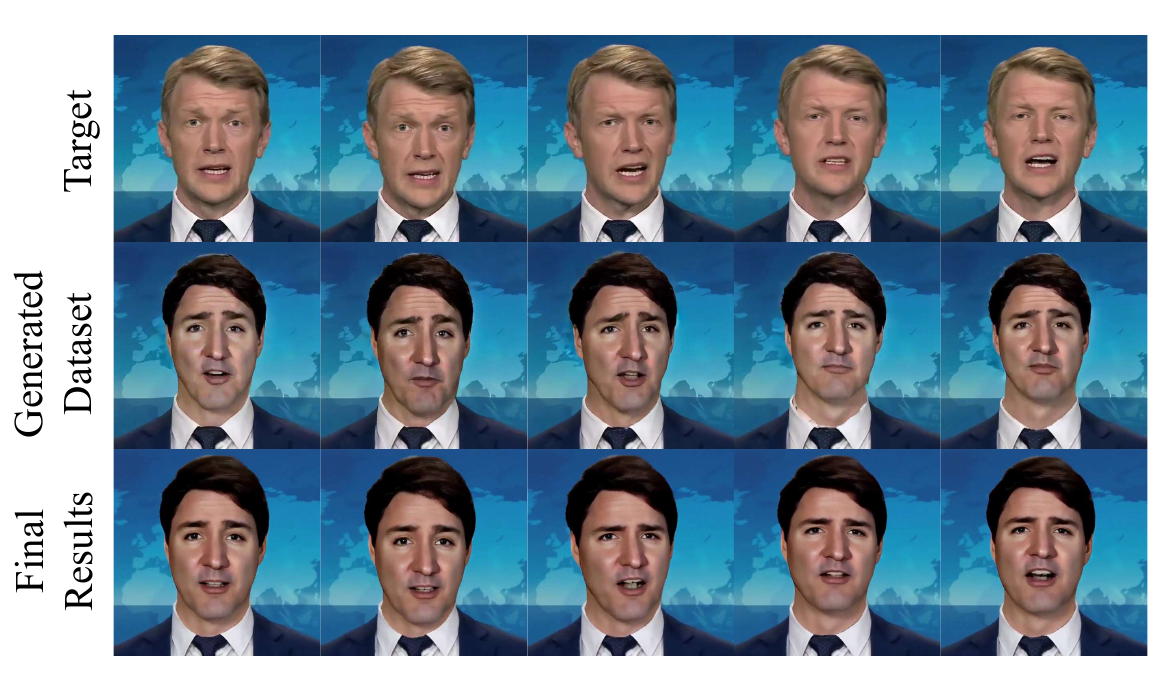}
  \caption{There exist some expression and pose misalignments between \textbf{TARGET} and \textbf{Generated Dataset}, and the \textbf{Generated Dataset} frames suffer from inconsistency.}
  \label{fig:gen}
  \vspace{-2mm}
\end{figure}
Due to the stochastic nature of diffusion model, the generated head swap frames Fig.~\ref{fig:gen} suffer from inconsistency. We re-extract SMPL-X parameters from the generated frames and train the neural Gaussian field on the retracked parameters.
\begin{figure}[t]
  \centering
  \includegraphics[width=0.95\linewidth]{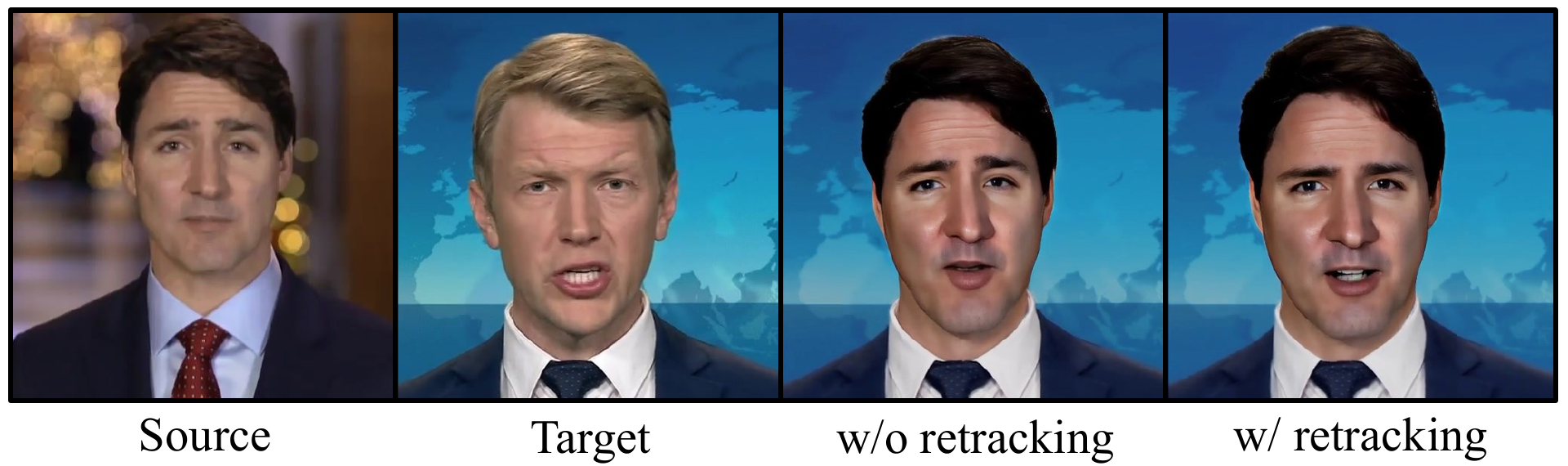}
  \caption{To address inconsistencies in head swap dataset frames, we retrack the data frames to mitigate these discrepancies. This approach effectively enhances the accuracy of facial expressions.}
  \label{fig:RT}
\end{figure}
As demonstrated in Fig.~\ref{fig:RT}, the inconsistency of generated frames leads to expression degeneration when omitting the retracking strategy. More vivid results can be seen in the supplementary video.

Besides, the temporal condition feature can handle the color inconsistency of generated datasets in some cases.

\subsubsection{Neural Rerendering}
\begin{figure}[htb]
  \centering
  \includegraphics[width=\linewidth]{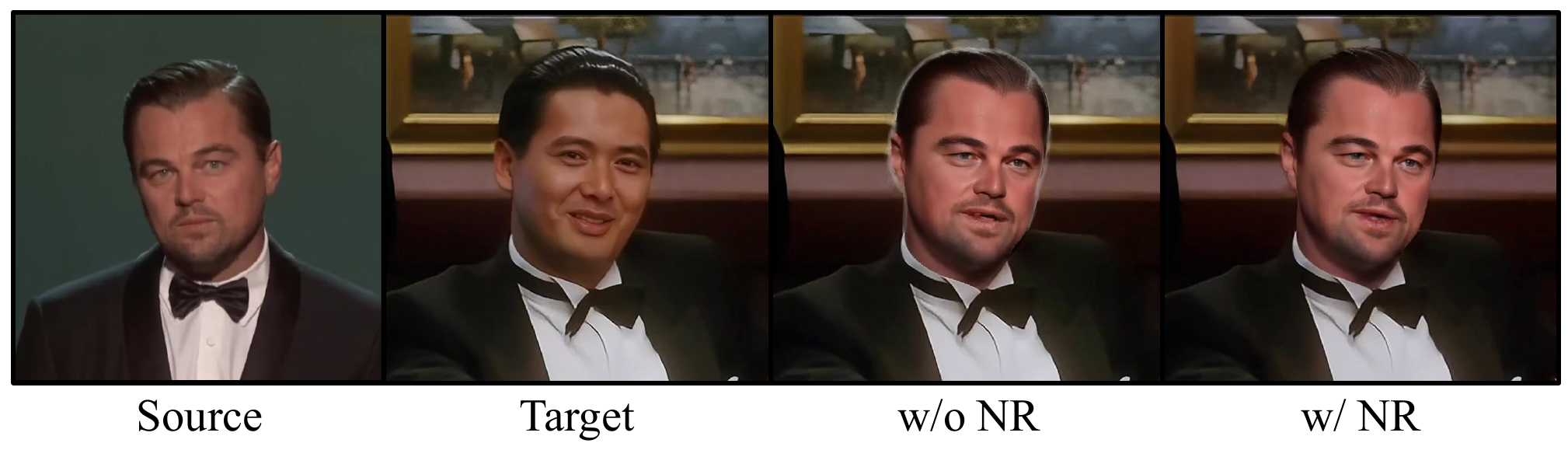}
  \caption{Neural Rerendering(NR) can seamlessly handle background mismatch.}
  \label{fig:NR}
  \vspace{-2mm}
\end{figure}
The huge mismatch in the head region between $\phi_{\textrm{fore}}$ and $I_{\textrm{tgt}}$ caused by the editing of head shapes and hairstyles makes the results unnatural. To address this issue, we developed a neural rerendering module that operates on the feature maps of both $\phi_{\textrm{fore}}$ and $I_{\textrm{tgt}}$. This module effectively handles the mismatch issues, ensuring a seamless integration of the head region with the background. As illustrated in Fig.~\ref{fig:NR}, our neural rerendering (NR) technique effectively merges the head region into the surrounding environment, enhancing the naturalness of the result. In contrast, directly blending in the RGB domain using a head mask fails to adequately address the mismatch problem.

% So we designed a neural rerendering module on the feature map of $\phi_{fore}$ and $I_{tgt}$ to handle mismatch issues naturally. As demonstrated in Fig.~\ref{fig:NR}, our neural rerendering (NR) can integrate the head region seamlessly into the background.

\subsubsection{Impact of K-Shot Fine-Tuning}
\begin{figure}[htb]
  \centering
  \includegraphics[width=0.95\linewidth]{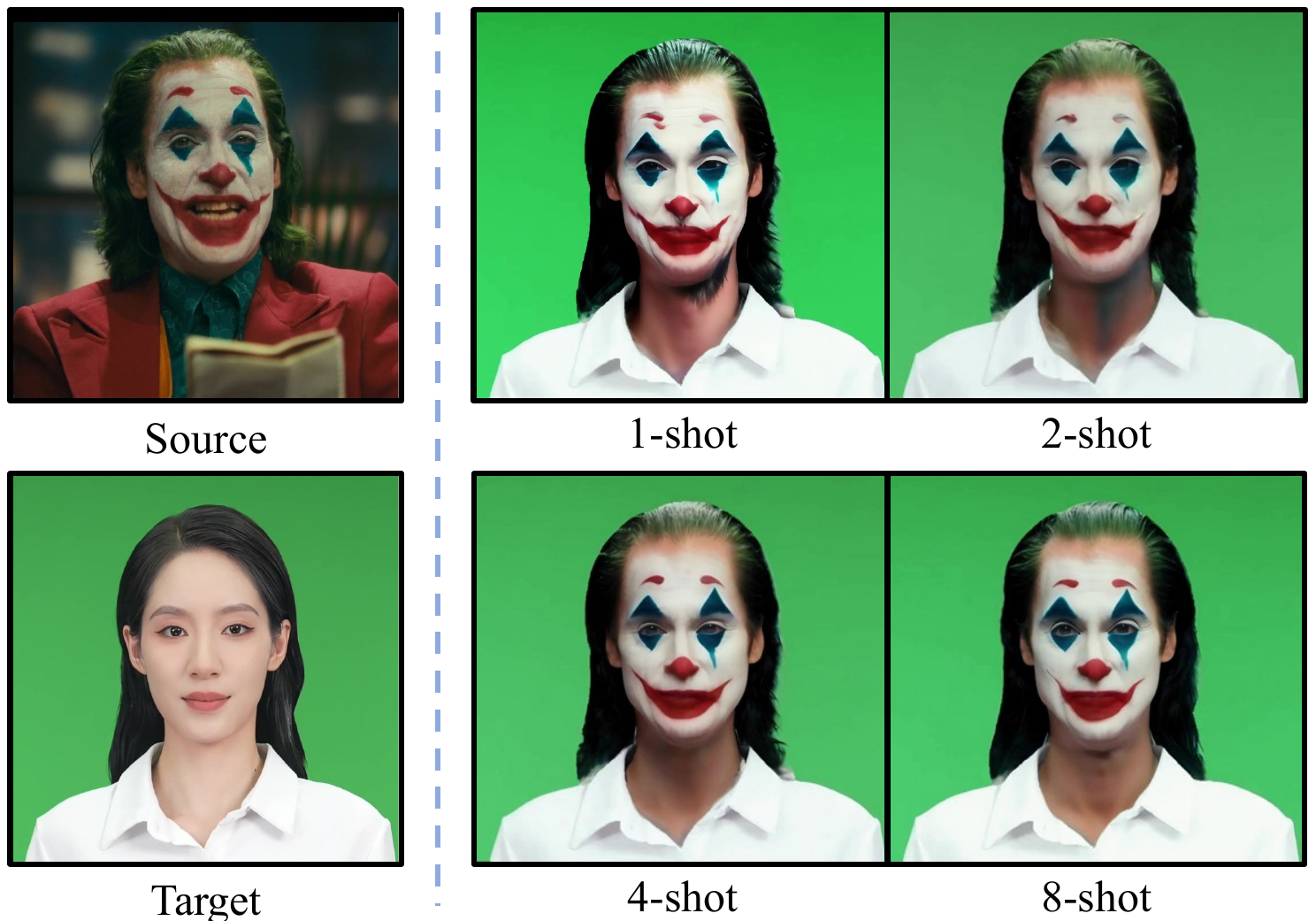}
  \caption{Our domain adaptation strategy accepts an arbitrary numbers of reference source images. We can achieve reasonable results with just one source frame as input. Increasing the number of source frames could further enhance identity preservation and facial details.}
  \label{fig:KSHOT}
  \vspace{-2mm}
\end{figure}
We further analyze the effect of decreasing the K-shot number with subject-specific fine-tuning. Specifically, our 2D portrait generation model takes 4-12 frames as input. We compare the results of different number of input images. The qualitative comparison results in Fig.~\ref{fig:KSHOT} demonstrate that 4 input frames can obtain identity-preserving results. One-shot input can also achieve the head swapping effect, however, it tends to introduce some artifacts and generally achieves lower identity similarity. Please refer to supplementary material for more detailed qualitative and vivid results.

\subsubsection{Neural Feature}
\begin{figure}[htb]
  \centering
  \includegraphics[width=\linewidth]{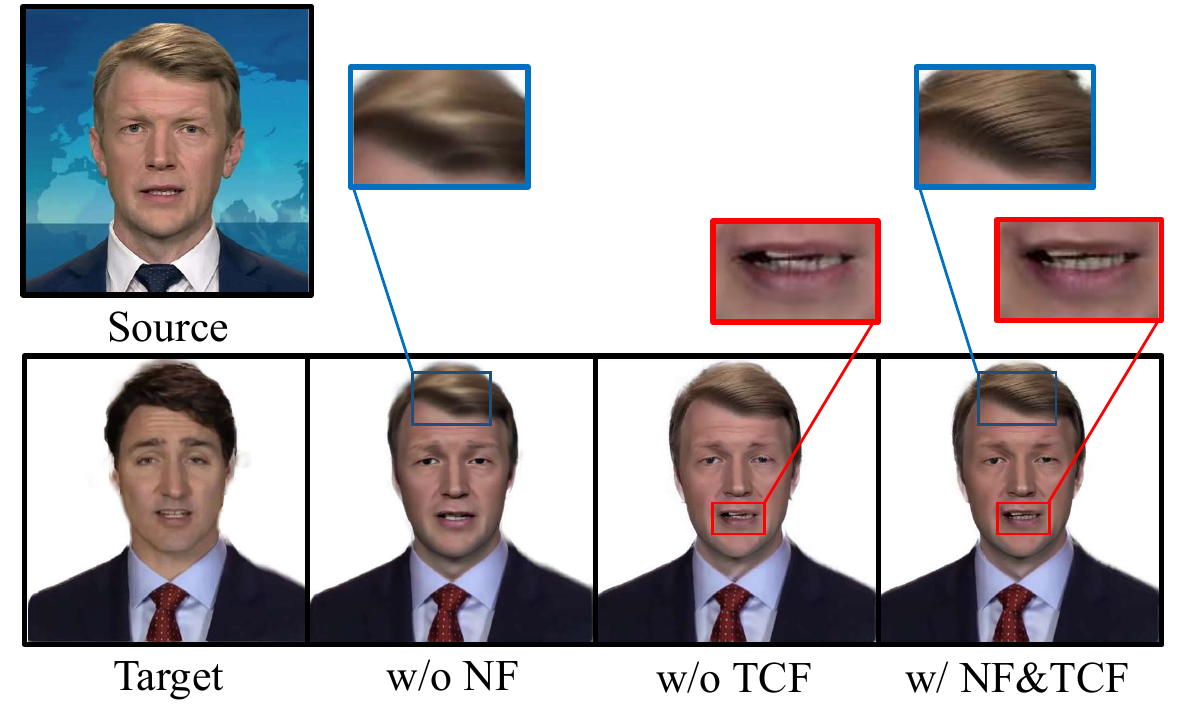}
  \caption{Lack of temporal conditioning features (TCF) leads to artifacts. Neural Gaussian Field (NF) performs better at representing high-frequency details}
  \label{fig:nf_tcf}
  \vspace{-2mm}
\end{figure}
While explicit 3D Gaussians with spherical harmonic (SH) coefficients per Gaussian offer efficient rendering, they struggle to faithfully reconstruct high-frequency details such as fine hair strands, subtle facial textures, and intricate lighting variations. This limitation arises because spherical harmonics, while effective for approximating low-frequency lighting, lack the expressive power to encode sharp, high-frequency features. As demonstrated in Fig.~\ref{fig:nf_tcf}, our proposed Neural Gaussian Field overcomes this issue by implicitly modeling high-frequency details through a learnable neural feature representation, rather than relying solely on explicit SH-based rendering.

\subsubsection{Temporal Condition Feature}
While diffusion models have demonstrated remarkable capabilities in generating high-quality images, their inherent stochastic sampling process introduces significant challenges for video-based head swapping. Due to the independent noise sampling at each timestep, the generated head-swap frames exhibit temporal inconsistencies, manifesting as flickering artifacts, unstable facial features, and unnatural expression transitions between frames. As clearly shown in Fig.~\ref{fig:nf_tcf}, this problem stems primarily from the absence of effective temporal conditioning mechanisms in standard diffusion approaches.

\subsubsection{Undress \& Paste Background}
\begin{figure}[htb]
  \centering
  \includegraphics[width=\linewidth]{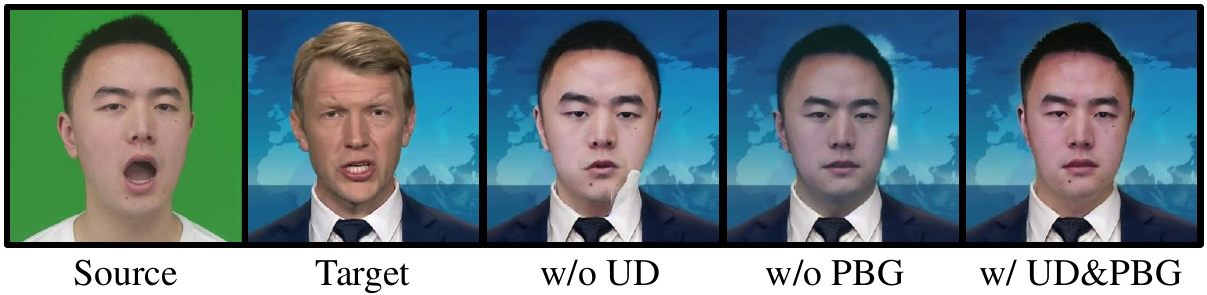}
  \caption{Undress (UD) and pasting background (PBG) can generate better head-swapped images.}
  \label{fig:pre}
  \vspace{-2mm}
\end{figure}

The clothes and background information in the source image are unnecessary and may lead to improper head-torso relationships in the generated dataset. As illustrated in Fig.~\ref{fig:pre}, our preprocessing module can generate head-swapped images with better head-torso alignment.
\section{Conclusion and Discussion}
% \subsection{Limitation}
 
% \subsection{Conclusion}
In this paper, we have introduced the video head swap system, GSwap. We adapted a pre-trained diffusion model to the source head domain and used it for inpainting on the target video to create a training dataset. The diffusion model's capability ensures high identity similarity with the source images. Additionally, we integrated a 3D Gaussian field onto the surface of SMPL-X to enhance the quality and ensure temporal consistency of the results. We also introduce a neural rerendering module that seamlessly blends the foreground head region with the background. Extensive experiments show that GSwap significantly improves quality compared to existing methods.

 The Sec.~\ref{arch-gen} presents a natural and high-fidelity image head-swapping method. Using this method, we generated the dataset for training portrait representation models. In fact, our head-swapping model can be adapted to other inpainting technique or image head-swapping methods.

 Our method relies on tracked SMPL-X and head segment, and thus large errors in tracking or segment may cause artifacts. As our method utilizes pretrained 2D portrait generation model for domain adaption, the bias and errors in these models may also influence the head swap results. We would like to clarify that the GSwap pipeline exhibits robustness across a wide range of scenarios. To verify this, we have included extreme cases featuring large head rotation angles and fast moving heads in the supplementary video. These examples visually demonstrate that even under such challenging conditions, our method maintains reliable SMPL-X tracking and segmentation, enabling the generation of high-quality dynamic portraits.

 \begin{figure}[htb]
  \centering
  \includegraphics[width=\linewidth]{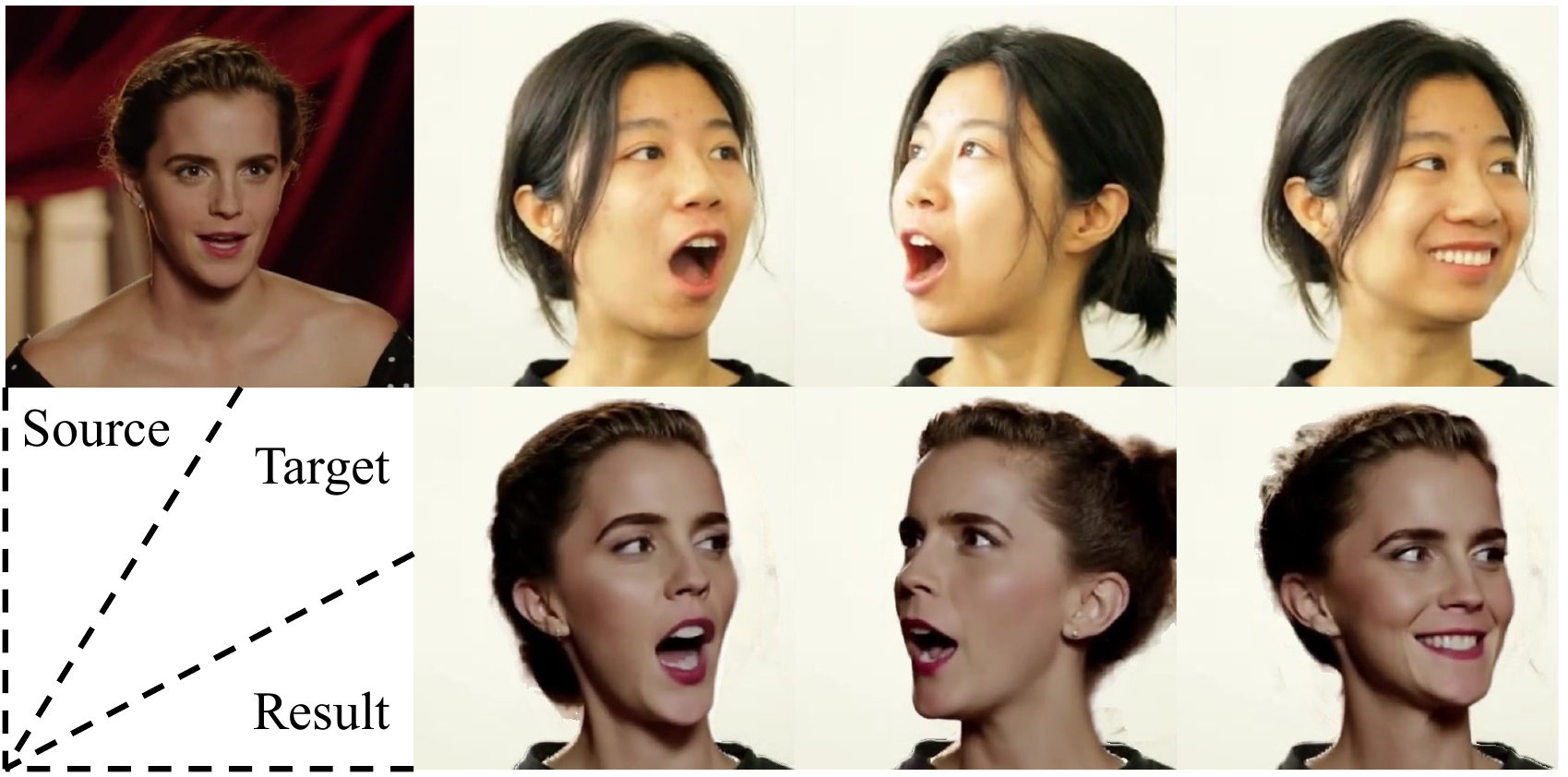}
  \caption{Extreme cases with large head rotation angles.}
  \label{fig:ec}
  \vspace{-2mm}
\end{figure}
 
\section{Ethics Statement}
Our video head swap system, GSwap, focuses on technical development. Our method can generate head swapping results with few shot images as input. Due to its ability to produce high-fidelity results and its high degree of flexibility in the generation process, misuse of our methods may raise ethical issues. Therefore, it is imperative to strictly prohibit any inappropriate behavior associated with its use. As a result, we require that the media data generated by our method clearly present itself as synthetic. Furthermore, we strongly believe that it is crucial to develop safeguarding measures to mitigate the potential for misuse.

\section{Acknownledgements}
This research was supported by the National Natural Science Foundation of China (No.62272433, No.62402468, No.U25A20390), and the Fundamental Research Funds for the Central Universities.

% \subsection{Discussion}
% \noindent\textbf{Ethics Statement.}
% Our video head swap system, SwapFrom4D, focuses on technical development. Our method
% can generate head swapping results with few shot images as input. Due to its ability to produce high-fidelity results and its high degree of flexibility in the generation process, misuse of our methods may raise ethical issues. Therefore, it is imperative to strictly 
% prohibit any inappropriate behavior associated with its use. As a result, we require that the media data generated by our method clearly present itself as synthetic. Furthermore, we strongly believe that it is crucial to develop safeguarding measures to mitigate the potential for misuse.

{
\bibliographystyle{IEEEtran}
\bibliography{egbib}
}
\begin{IEEEbiography}[{\includegraphics[width=1in,height=1.25in,clip,keepaspectratio]{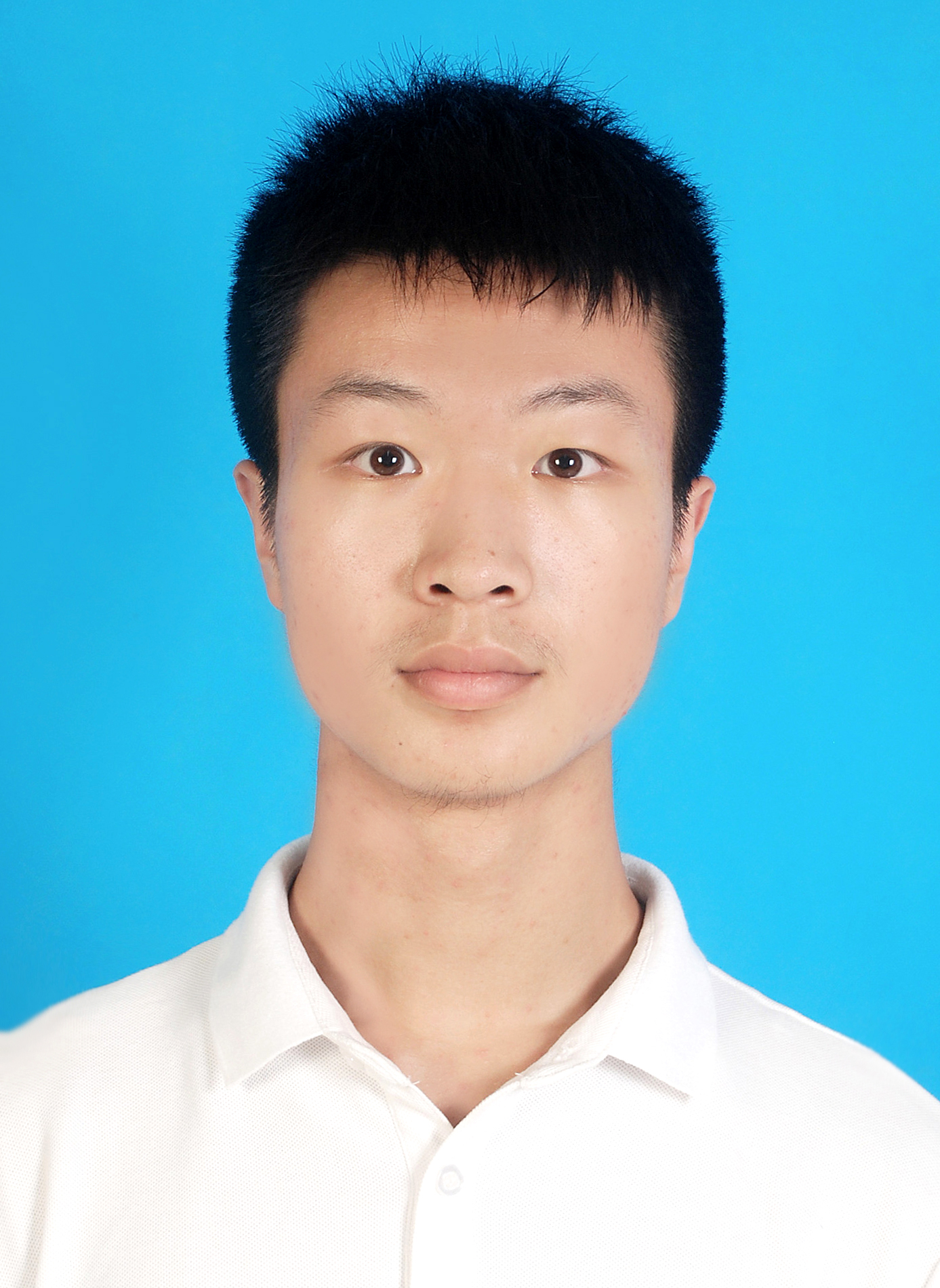}}]{Jingtao Zhou}
is a joint Ph.D. student at the School of Mathematical Sciences, University of Science and Technology of China and the Department of Computer Science, City University of Hong Kong. His research interests include computer vision and deep learning.
\end{IEEEbiography}
\vspace{20mm}
\begin{IEEEbiography}[{\includegraphics[width=1in,height=1.25in,clip,keepaspectratio]{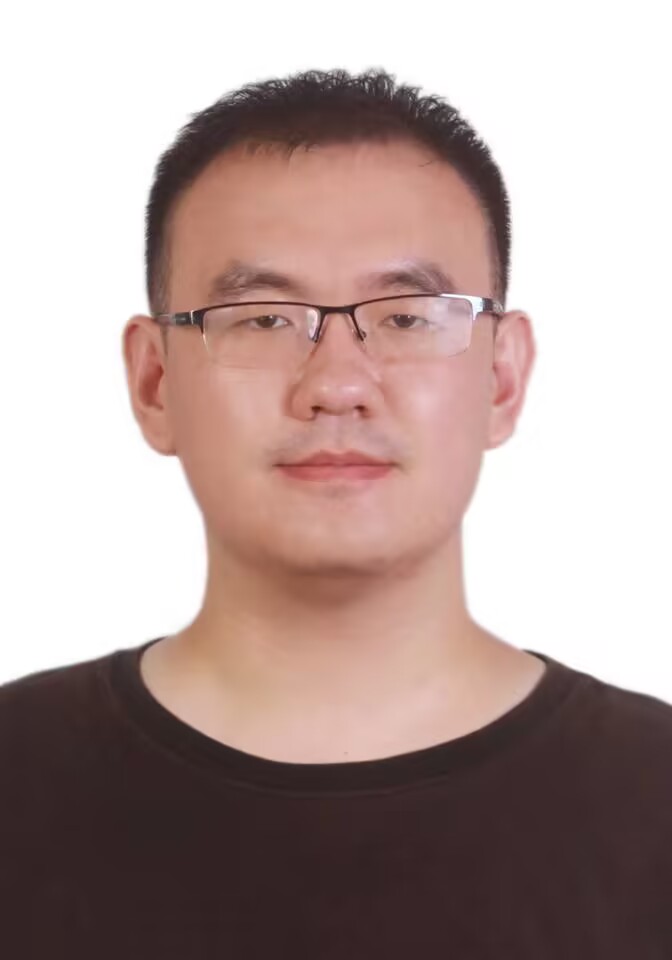}}]{Xuan Gao}
is a Ph.D. student at the School of Mathematical Sciences, University of Science and Technology of China. His research interests include computer graphics and computer vision.
\end{IEEEbiography}
\vspace{20mm}
\begin{IEEEbiography}[{\includegraphics[width=1in,height=1.25in,clip,keepaspectratio]{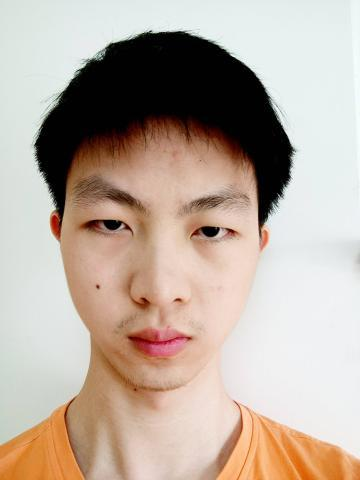}}]{Dongyu Liu}
is a master student at the School of Mathematica Sciences, University of Science and Technology of China. His research interests include computer graphics and computer vision.
\end{IEEEbiography}
\vspace{-5mm}
\begin{IEEEbiography}[{\includegraphics[width=1in,height=1.25in,clip,keepaspectratio]{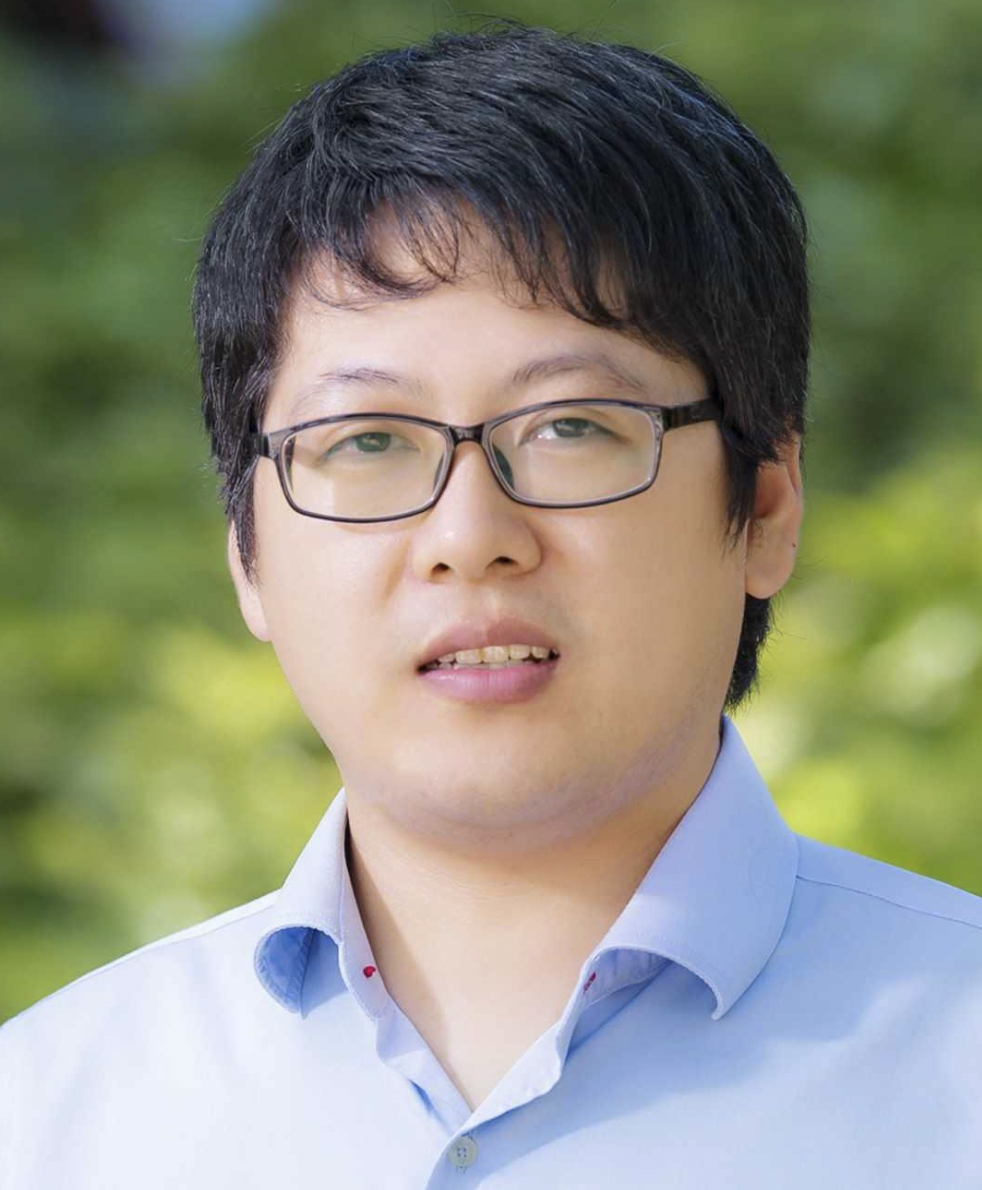}}]{Junhui Hou} (Senior Member, IEEE) is an Associate Professor with the Department of Computer Science, City University of Hong Kong. His research interests are multi-dimensional visual computing. Dr. Hou received the Early Career Award (3/381) from the Hong Kong Research Grants Council in 2018, the NSFC Excellent Young Scientists Fund in 2024, and IEEE SPS Best Paper Award in 2025. He is serving as a Senior Area Editor for IEEE Trans. Image Processing, and an Associate Editor for IEEE Trans. on Visualization and Computer Graphics and IEEE Trans. on Multimedia.
\end{IEEEbiography}
\vspace{-5mm}
\begin{IEEEbiography}[{\includegraphics[width=1in,height=1.25in,clip,keepaspectratio]{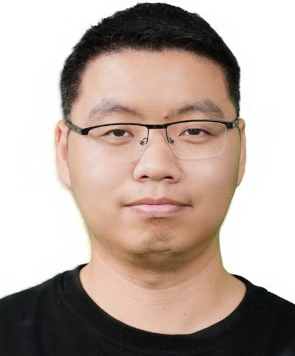}}]{Yudong Guo} is an assistant professor at University of Science and Technology of China (USTC). He got the Ph.D. degree from USTC in 2021, supervised by Prof. Juyong Zhang. Before that, He received bachelor degree in Statistics in 2015 from USTC. From fall 2016 to spring 2017, He was a research assistant in the MultiMedia Lab at Nanyang Technological University under supervision of Prof. Jianfei Cai and Prof. Jianmin Zheng. His research interests include 3D vision and digital human.
\end{IEEEbiography}
\vspace{-5mm}
\begin{IEEEbiography}[{\includegraphics[width=1in,height=1.25in,clip,keepaspectratio]{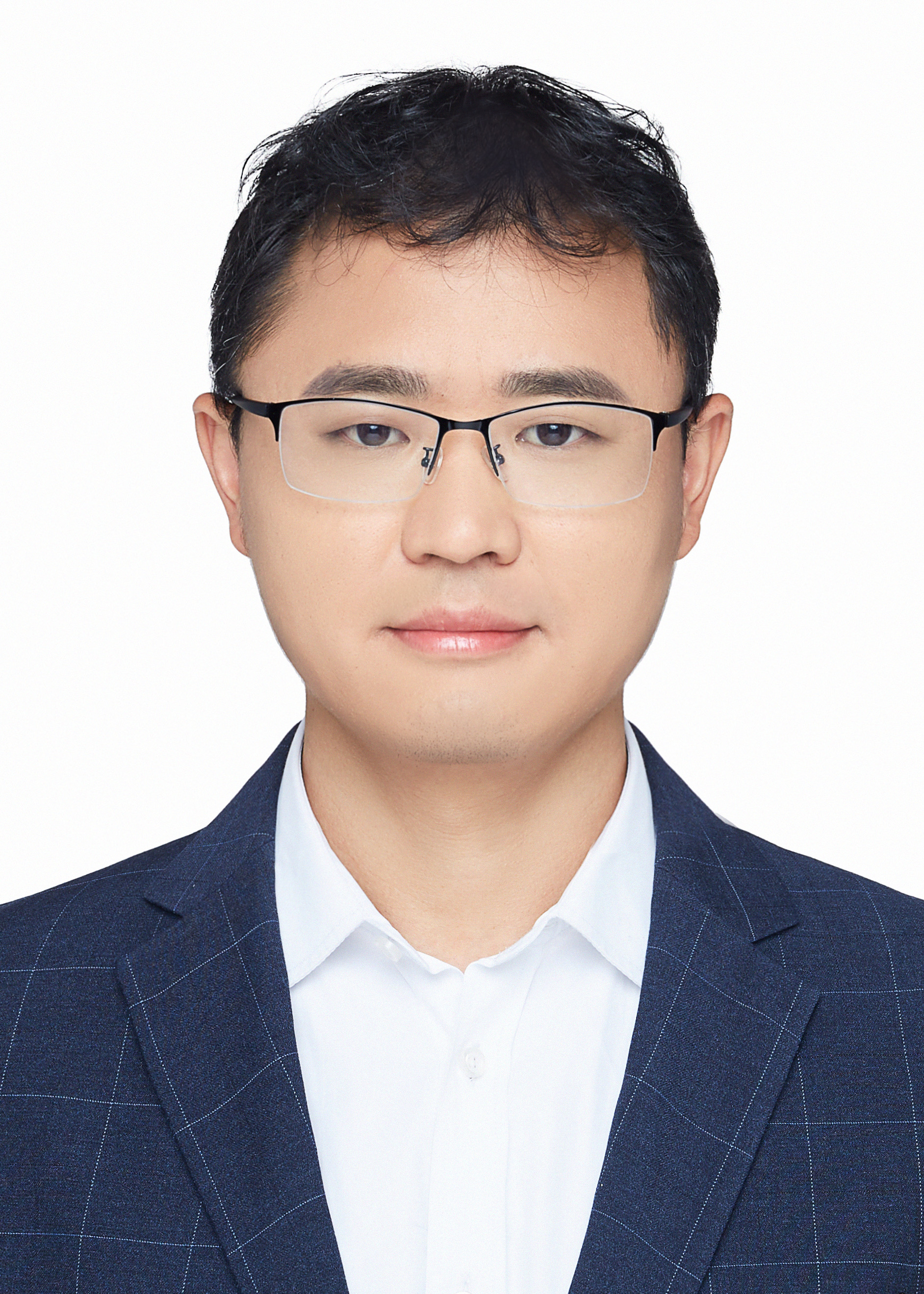}}]{Juyong Zhang} is a professor in the School of Mathematical Sciences at University of Science and Technology of China. He received the BS degree from the University of Science and Technology of China in 2006, and the Ph.D degree from Nanyang Technological University, Singapore. He mainly conducts research at the intersection of Vision, Graphics, and AI with a special focus on capturing, modeling and synthesizing objects, humans and large-scale scenes. He is an associate editor of IEEE Transactions on Multimedia and IEEE Computer Graphics and Applications.
\end{IEEEbiography}

\end{document}